\documentclass[pdflatex,sn-nature]{sn-jnl}

\usepackage{graphicx}%
\usepackage{multirow}%
\usepackage{amsmath,amssymb,amsfonts}%
\usepackage{amsthm}%
\usepackage{mathrsfs}%
\usepackage[title]{appendix}%
\usepackage{xcolor}%
\usepackage{textcomp}%
\usepackage{manyfoot}%
\usepackage{booktabs}%
\usepackage{algorithm}%
\usepackage{algorithmicx}%
\usepackage{algpseudocode}%
\usepackage{listings}%

\usepackage{tabularx}
\usepackage{siunitx}

\usepackage{forest}
\usepackage{amssymb}

\theoremstyle{thmstyleone}%
\theoremstyle{thmstyletwo}%

\theoremstyle{thmstylethree}%

\begin{document}

\title[Clinical Information Retrieval from Finnish EHRs]{Automating Clinical Information Retrieval from Finnish Electronic Health Records Using Large Language Models}


\author*[1]{\fnm{Mikko} \sur{Saukkoriipi}}\email{mikko.1.saukkoriipi@aalto.fi}
\author[2]{\fnm{Nicole} \sur{Hernandez}}
\author[1]{\fnm{Jaakko} \sur{Sahlsten}}
\author[1]{\fnm{Kimmo} \sur{Kaski}}
\author[2,3]{\fnm{Otso} \sur{Arponen}}

\affil[1]{\orgdiv{Department of Computer Science}, \orgname{Aalto University School of Science}, \orgaddress{\city{Espoo}, \postcode{02150}, \country{Finland}}}
\affil[2]{\orgdiv{Faculty of Medicine and Health Technology}, \orgname{Tampere University}, \orgaddress{\city{Tampere}, \postcode{33520}, \country{Finland}}}
\affil[3]{\orgdiv{Department of Oncology, TAYS Cancer Centrer}, \orgname{Tampere University Hospital}, \orgaddress{\city{Tampere}, \postcode{33520}, \country{Finland}}}

\abstract{Clinicians often need to search for patient-specific information within electronic health records (EHRs) containing heterogeneous longitudinal documentation, a task that can be time-consuming, cognitively demanding, and error-prone. Here we present a locally deployable Clinical Contextual Question Answering (CCQA) framework that answers patient-specific clinical questions using information from EHRs without transmitting data outside institutional boundaries. Open-source large language models (LLMs) ranging from 4 billion (4B) to 70 billion (70B) parameters were benchmarked under fully offline conditions. The models were evaluated using 1,664 expert-annotated question-answer pairs derived from records of 183 patients undergoing breast cancer screening, diagnosis, or treatment. The dataset contained predominantly Finnish clinical text with occasional English and Latin terminology. In free-text generation, Llama-3.1-70B achieved 95.3\% accuracy and 97.3\% consistency in semantically equivalent question variants, while the mid-sized Qwen3-30B-A3B-2507 model achieved comparable performance despite its smaller parameter count. In a multiple-choice setting, the models achieved similar accuracy while enabling probabilistic calibration analysis, which revealed considerable variability in calibration between models. This indicates that for clinical deployment, calibration should be considered in conjunction with accuracy. We also found that low-precision quantization (4-bit and 8-bit) preserved predictive performance for most models while markedly reducing GPU memory requirements, improving feasibility for on-site hospital deployment. Models adapted to medical tasks or to the Finnish language did not demonstrate systematic advantages over general-purpose LLMs. In addition, clinical evaluation of responses from the best-performing model indicated that 2.9\% of outputs contained clinically significant errors. Semantically equivalent question formulations produced discordant responses in a small proportion of cases, including instances in which one formulation yielded a correct response and the other a clinically significant error (0.96\% of all responses). These findings indicate that locally hosted open-source LLMs can accurately retrieve clinically relevant patient-specific information from longitudinal EHRs using natural-language queries, supporting secure clinical information retrieval while emphasizing the need for careful validation, robustness assessment, and human oversight in clinical deployment.}





\maketitle


\section{Introduction}\label{introduction}

\begin{figure*}[t]
\centering
\includegraphics[width=0.98\textwidth]{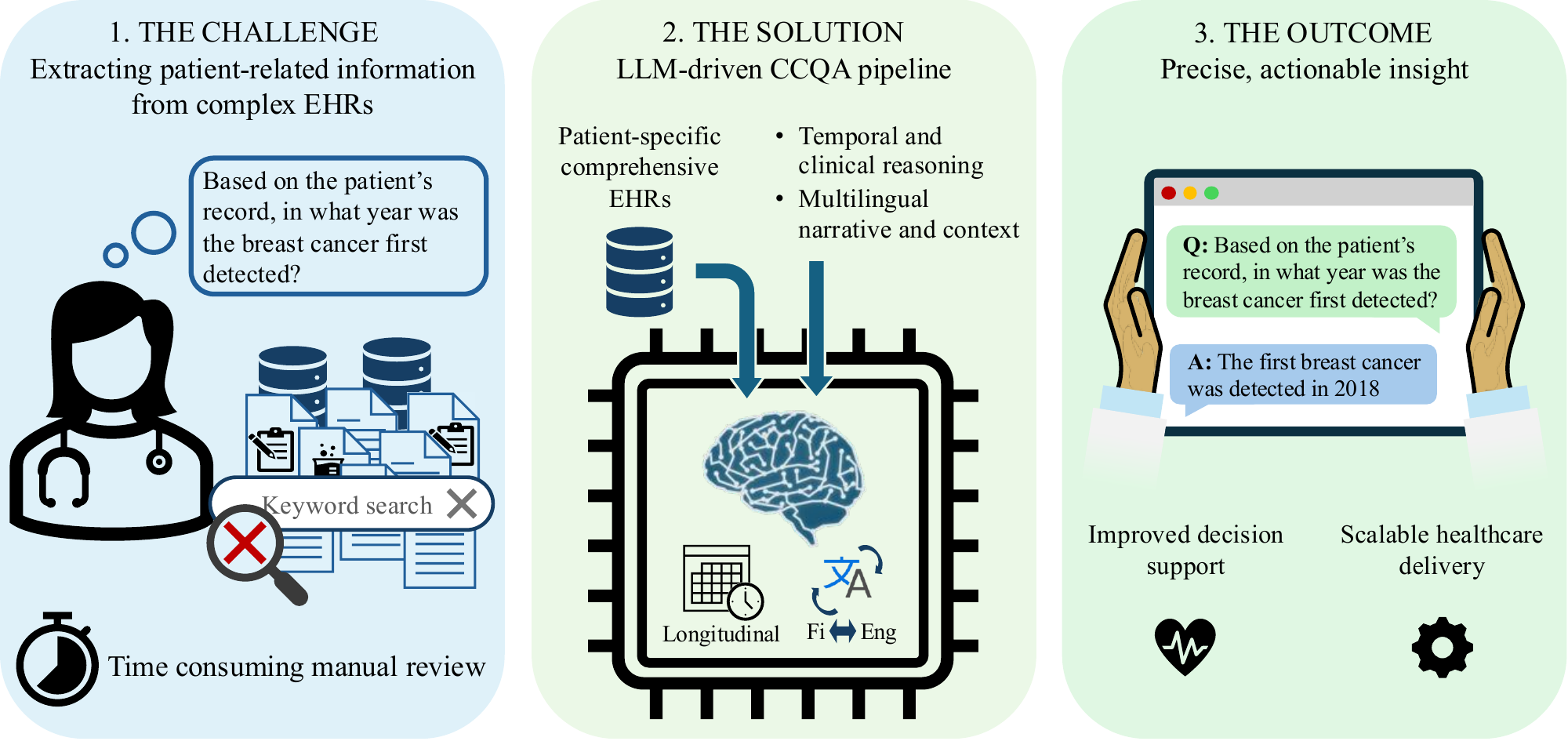}
\caption{Visual abstract of the Clinical Contextual Question Answering (CCQA) pipeline. A clinician submits a natural language question about a patient record. A long-context large language model processes the complete electronic health record and generates a context-aware answer, which is returned to the clinician for use in clinical practice.}
\label{fig:pipeline}
\end{figure*}

Modern clinical practice faces an information paradox. Electronic health records (EHRs) contain extensive longitudinal patient data, yet the volume and fragmentation of these records impose a substantial cognitive burden on clinicians in their daily practice \cite{beasley2011information, arndt2017tethered, mazur2019association}. In oncology and other multidisciplinary medical specialities, patient histories are distributed across years of heterogeneous EHR documentation, including radiology and pathology reports, surgical summaries, and medication records \cite{hanauer2015supporting}. Synthesizing this information during active consultations is time-consuming and prone to oversight, potentially increasing workload, and delaying interpretation, particularly when clinicians rely on manual review or keyword-based search across dispersed documentation \cite{natarajan2010analysis}. Systems must therefore support flexible, natural language querying of patient EHRs and return precise, context-aware answers grounded in complete longitudinal records. Such capabilities could improve workflow efficiency, reduce cognitive burden, and support higher-quality clinical decision-making \cite{beasley2011information, wiest2024privacy}.

Large language models (LLMs) have recently been applied to question answering over EHRs to improve access to complex clinical information~\cite{kothari2025question}. In this work, we refer to patient-specific QA grounded in longitudinal clinical records as clinical contextual question answering (CCQA). Unlike standard document retrieval, this task requires synthesizing information across multiple sources to derive clinically relevant variables such as diagnosis timing, disease characteristics, or treatment sequences. Previous work has shown that LLMs can encode clinical knowledge and achieve strong performance on medical question answering tasks~\cite{singhal2023large, thirunavukarasu2023large, 2025medgemma}, and these models have been applied to EHR-based QA using synthetic patient records~\cite{kothari2025question}. However, synthetic data may not capture the scale and complexity of real-world clinical information, which spans years of heterogeneous structured and unstructured patient records. As a result, patient-specific QA over EHRs using LLMs remains comparatively underexplored in studies using real-world clinical data.

This gap is further pronounced in healthcare settings involving non-English clinical texts~\cite{ong2026large}. In Finland’s healthcare system, EHRs are primarily written in Finnish, with embedded English and Latin terminology, domain-specific abbreviations, and complex linguistic structures that challenge general-domain language models~\cite{virtanen2019multilingual, myllyla2025extracting}. Prior work in Finnish clinical NLP is limited and has focused mainly on task-specific classification and information extraction, with no prior studies, to our knowledge, evaluating generative LLMs for patient-specific clinical question answering over Finnish-language EHRs~\cite{virtanen2019multilingual, jaskari2024dr, myllyla2025extracting}. More broadly, these challenges reflect pretraining imbalances, as medical text constitutes only a small fraction of large-scale corpora and many languages remain under-represented~\cite{ong2026large, gururangan2020don, 2025medgemma}. Together, this setting provides a stringent test of model generalization beyond standard benchmarks.

Historically, many clinically relevant natural language processing systems have relied on retrieval-augmented generation (RAG) to manage long documents by retrieving and processing subsets of the record~\cite{lewis2020retrieval, kothari2025question}. Such approaches can be limited in tasks requiring longitudinal reasoning, particularly when it is unclear which parts of the patient history are relevant~\cite{wiest2024privacy, shickel2017deep}. In complex specialties such as oncology, accurate interpretation often requires integrating events separated by months or years, distinguishing active disease from historical or resolved findings, and reasoning over the clinical timeline~\cite{zhang2018patient2vec}. Recent advances in large language models have enabled substantially longer input sequences~\cite{dao2022flashattention, beltagy2020longformer, su2024roformer}, allowing models to process full patient EHRs and support end-to-end reasoning over longitudinal clinical histories.

Despite rapid progress in LLM development, several important questions remain unresolved~\cite{kaplan2020scaling}. First, it is unclear how model scale, architecture, and domain specialization influence performance in real-world CCQA over longitudinal EHRs~\cite{bommasani2022opportunitiesrisksfoundationmodels, thirunavukarasu2023large}. Second, although low-precision quantization has been proposed to enable deployment on resource-constrained hardware, its impact on clinical reliability remains unknown~\cite{hoffmann2022training}. Third, most prior medical LLM research focuses on general question answering, narrowly defined extraction tasks, or synthetic data, leaving uncertainty about performance in multilingual, longitudinal clinical narratives~\cite{safavi2025benchmarking, gallifant2025evaluating}. Finally, accuracy alone is insufficient for clinical adoption. Consistency across semantically equivalent queries and calibration of model confidence are essential for trustworthy integration into clinical workflows~\cite{kadavath2022language, ghassemi2020review, thirunavukarasu2023large, nakkiran2025trained}.

Recent advances enable language models to process complete patient records while remaining deployable on hospital-governed hardware, supporting privacy-preserving analysis of sensitive clinical data~\cite{wiest2024privacy, dennstadt2025implementing, kothari2025question, 2025medgemma}. However, empirical evaluation remains limited in real-world multilingual oncology settings involving longitudinal EHRs~\cite{safavi2025benchmarking, gallifant2025evaluating}.

In this study, we present a systematic benchmark of retrieval-free long-context LLMs for Clinical Contextual Question Answering in a multilingual oncology environment. Using a dataset of 1,664 expert-annotated question–answer pairs derived from EHRs of 183 patients undergoing breast cancer screening or treatment, we evaluated five families of open-source models ranging from 4 billion to 70 billion parameters. Our analysis examines accuracy, response consistency across semantically equivalent queries, calibration of model confidence, memory footprint under extended sequence lengths, and the impact of quantization and prompting strategies. In addition, we conducted a clinical safety evaluation of incorrect responses from the highest-performing model to assess their clinical significance. By providing a comprehensive comparison of locally deployable models operating on full patient histories, this work demonstrates the feasibility of privacy-preserving, retrieval-free clinical information retrieval and offers practical insights into how model scale, architecture, and deployment constraints shape real-world clinical AI performance.

\section{Results}
\label{results}

We evaluated 16 instruction-tuned large language models from five developers on 1,664 clinical question–answer pairs derived from electronic health records of 183 patients undergoing breast cancer screening or treatment. Two models exceeded the available 160 GB GPU memory limit and were thus excluded, leaving 14 models for evaluation against a rule-based approach that served as the baseline result.

\begin{table*}[ht]
\centering
\footnotesize
\caption{\textbf{Clinical contextual question answering performance and deployment metrics.} Accuracy is reported for two question variants (V1 and V2) and their consistency across variants. Peak VRAM and latency are reported as the minimum and maximum per-query values across the dataset. The best results per column are shown in bold. OOM indicates models exceeding the 160\,GB memory limit. Models marked with $^{\dagger}$ were evaluated on a reduced subset ($N=955$) due to context window constraints. All other models were evaluated on the full dataset ($N=1{,}664$).}
\label{tab:main_results}
\resizebox{\textwidth}{!}{  
\begin{tabular}{llccccccc}
\toprule
\multirow{2}{*}{\textbf{Developer}} &
\multirow{2}{*}{\textbf{Model}} &
\multicolumn{3}{c}{\textbf{Accuracy (\%) $\uparrow$}} &
\multicolumn{2}{c}{\textbf{VRAM (GB) $\downarrow$}} &
\multicolumn{2}{c}{\textbf{Latency (s) $\downarrow$}} \\
\cmidrule(lr){3-5}\cmidrule(lr){6-7}\cmidrule(lr){8-9}
 &  & \textbf{V1} & \textbf{V2} & \textbf{Cons.} &
\textbf{Min} & \textbf{Max} &
\textbf{Min} & \textbf{Max} \\
\midrule
\multirow{4}{*}{Google}
 & Gemma-3-4B-it~\cite{2025gemma3} & 72.6 & 68.2 & 83.2 & 8.5 & 18.4 & 0.2 & 5.8 \\
 & Gemma-3-27B-it~\cite{2025gemma3} & 92.7 & 92.8 & 96.3 & 52.4 & 71.9 & 0.7 & 16.6 \\
 & MedGemma-4B-it~\cite{2025medgemma} & 74.5 & 73.1 & 89.6 & 8.5 & 18.4 & 0.3 & 5.9 \\
 & MedGemma-27B-text-it~\cite{2025medgemma} & 92.1 & 90.0 & 94.0 & 51.6 & 71.2 & 0.7 & 16.7 \\
\addlinespace
\multirow{3}{*}{DeepSeek}
 & R1-Distill-Qwen-7B~\cite{2025deepseek-r1} & 50.4 & 51.2 & 43.4 & 14.7 & 22.6 & 5.7 & 41.4 \\
 & R1-Distill-Qwen-32B~\cite{2025deepseek-r1} & 92.7 & 94.6 & 94.7 & 62.1 & 77.7 & 11.0 & 91.3 \\
 & R1-Distill-Llama-70B~\cite{2025deepseek-r1} & 94.8 & \textbf{95.1} & 97.0 & 132.8 & 148.2 & 18.7 & 153.3 \\
\addlinespace
\multirow{4}{*}{Meta}
 & Llama-3.1-8B~\cite{2024llama3} & 79.9 & 80.2 & 94.2 & 15.5 & 22.7 & 0.2 & 2.7 \\
 & Llama-3.1-70B~\cite{2024llama3} & \textbf{95.3} & 94.2 & 97.3 & 132.5 & 148.2 & 1.2 & 19.7 \\
 & Llama-3.3-70B~\cite{2024llama3} & 94.9 & 94.9 & 97.9 & 132.5 & 148.2 & 1.2 & 19.8 \\
 & Llama-4-Scout-17B~\cite{MetaAI2025Llama4} & OOM & OOM & -- & -- & -- & -- & -- \\
\addlinespace
OpenAI & GPT-OSS-20B~\cite{openai2025gptoss120bgptoss20bmodel} & OOM & OOM & -- & -- & -- & -- & -- \\
\addlinespace
\multirow{2}{*}{Qwen}
 & Qwen3-4B-2507~\cite{yang2025qwen3} & 91.1 & 89.8 & 95.3 & \textbf{8.0} & \textbf{15.0} & 0.2 & 2.7 \\
 & Qwen3-30B-A3B-2507~\cite{yang2025qwen3} & 95.1 & 94.4 & 97.7 & 57.2 & 61.8 & 2.7 & 8.9 \\
\addlinespace
\multirow{2}{*}{LumiOpen}
 & Llama-Poro-2-8B~\cite{poro2_2025}$^{\dagger}$ & 70.2 & 69.3 & 82.1 & 15.3 & 16.8 & 0.3 & 19.5 \\
 & Llama-Poro-2-70B~\cite{poro2_2025}$^{\dagger}$ & 90.6 & 91.8 & 92.5 & 132.1 & 135.1 & 2.1 & 154.1 \\
 \addlinespace
 \midrule
Baseline & Rule-based (substring) & 45.4 & 45.4 & \textbf{100} & -- & -- & \textbf{$<$0.1} & \textbf{$<$0.1} \\
\bottomrule
\end{tabular}
}%
\end{table*}

\subsection*{Overall model accuracy}
The accuracy of the models varied substantially across architectures (Table~\ref{tab:main_results}). For the primary question formulation (V1), the highest performance was achieved by larger models, including Llama-3.1-70B (95.3\%), Llama-3.3-70B (94.9\%), DeepSeek-R1-Distill-Llama-70B (94.8\%), and Qwen3-30B-A3B-2507 (95.1\%). The 30B-parameter Qwen3 model achieved accuracy comparable to that of the best 70B-parameter models. Performance also varied among models of similar scale. Qwen3-4B-2507 achieved an accuracy of 91.1\%, whereas the reasoning-oriented R1-Distill-Qwen-7B achieved only 50.4\% accuracy. Performance generally increased with model scale across evaluated model families. For example, within the DeepSeek R1-Distill family, accuracy increased with scale from 50.4\% to 92.7\% and 94.8\% for the 7B, 32B, and 70B models, respectively. The rule-based substring baseline achieved substantially lower accuracy (45.4\%), indicating that many answers required contextual interpretation beyond simple lexical matching.

The confidence intervals for model accuracy are reported in Supplementary Table~5. Statistical comparisons using paired McNemar tests indicated that Llama-3.1-70B significantly outperformed Gemma-3-27B and MedGemma-27B (Holm-adjusted \(P < 0.001\)), whereas differences among the top-performing models were not statistically significant (Supplementary Table~6).

\subsection*{Robustness and consistency of responses}
The robustness to linguistic variation was evaluated using two semantically equivalent question formulations: a primary formulation (V1) and a rephrased variant (V2) used to assess consistency (Table~\ref{tab:main_results}). Accuracy on the rephrased questions (V2) closely matched that of the primary formulation (V1) across most models. Differences between formulations showed no consistent direction, as some models performed better with V1 and others with V2.

Consistency was defined as agreement in the semantic meaning of model outputs across the two formulations, irrespective of correctness. Consistency was highest among the top-performing models, with Llama-3.3-70B achieving 97.9\%. Qwen3-30B-A3B-2507 and Llama-3.1-70B also showed high consistency (97.7\% and 97.3\%, respectively). In contrast, R1-Distill-Qwen-7B showed substantially lower consistency (43.4\%).

\subsection*{Domain- and language-specialized models}
Medical domain–adapted models (MedGemma) and Finnish language–specialized models (Llama-Poro-2) did not demonstrate systematic performance advantages over generalist instruction-tuned models (Table~\ref{tab:main_results}). At comparable parameter scales, their accuracy was generally comparable to but not consistently higher than that of generalist models. For example, MedGemma models achieved similar accuracy to Gemma models of the same size, while Finnish-language Llama-Poro-2 models showed lower accuracy than other Llama-family models evaluated on the same tasks, although these models were evaluated on a reduced dataset due to context window constraints. These results indicate that domain-specific pretraining or language specialization alone does not guarantee improved performance for clinical question answering on Finnish-language electronic health records.

\subsection*{Reasoning models}
DeepSeek R1-Distill models showed accuracy comparable to similarly sized generalist models at larger parameter scales (Table~\ref{tab:main_results}). R1-Distill-Llama-70B achieved 94.8\% accuracy, similar to Llama-3.1-70B (95.3\%) and Llama-3.3-70B (94.9\%). The median reasoning trace length for R1-Distill-Llama-70B was 295 tokens (range 135--1001 tokens). In contrast, R1-Distill-Qwen-7B showed lower accuracy (50.4\%) and consistency (43.4\%) than non-reasoning models of comparable size.

\subsection*{Performance variability across clinical questions}
The performance of the model varied substantially between individual clinical questions despite similar aggregate scores (Supplementary Tables~2 and~3). The most variable question (CQ6), which assessed whether neoadjuvant therapy had been administered prior to surgery, showed accuracies ranging from 4.7\% to 98.7\% across systems. Another challenging question was CQ14, which required the extraction of the percentage of progesterone receptor expression and showed accuracies ranging from 25.0\% to 95.7\%. The mean accuracy between models ranged from 62.8\% (CQ6) to 93.7\% (CQ3). We found that seven questions consistently showed high accuracy in all models, with mean accuracy ranging from 88.9\% to 93.7\%. The highest accuracy per-question was distributed between multiple models and no single model achieved the best performance on all questions. 

\begin{figure}[ht]
\centering
\includegraphics[width=\textwidth]{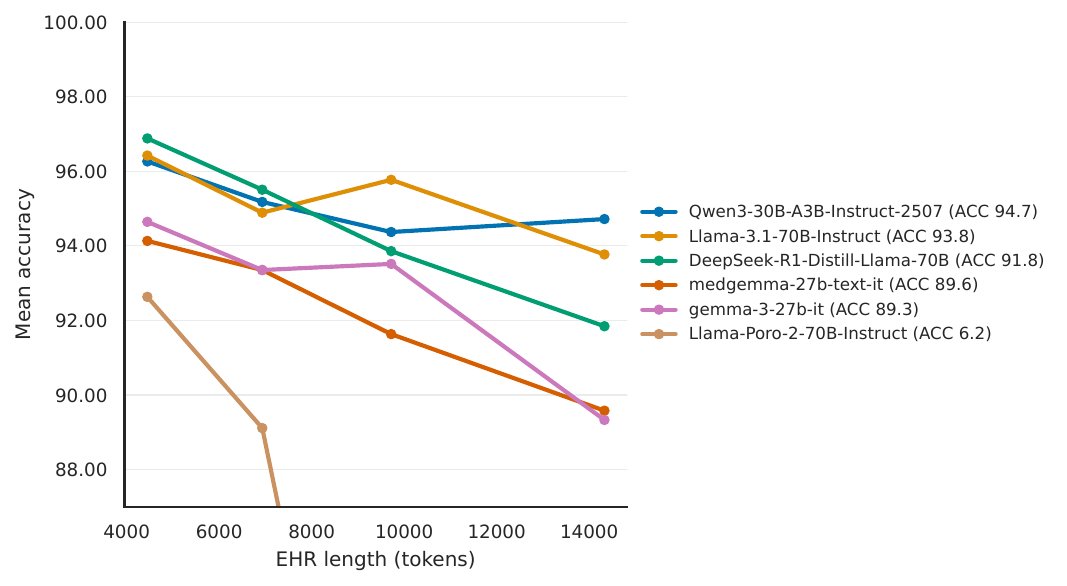}
\caption{Accuracy as a function of EHR length for the best-performing model from each model family. Results are shown across quartiles of EHR length (Q1–Q4), each containing 25\% of the dataset. The x-axis represents the median EHR length (tokens) within each quartile, and the y-axis shows average accuracy. Accuracy values reported next to each model correspond to performance in the longest-length quartile (Q4).}
\label{fig:accuracy_vs_ehr_length}
\end{figure}

\subsection*{Performance as a function of EHR length}
Accuracy varied with EHR length (Fig.~\ref{fig:accuracy_vs_ehr_length} and Supplementary Table~4). Records were stratified into quartiles spanning 1{,}420–26{,}739 tokens (Q1–Q4, shortest to longest). Several models maintained high accuracy across quartiles with only modest declines, including Llama-3.1-70B (96.4\% to 93.8\%), Llama-3.3-70B (96.6\% to 92.6\%), and Qwen3-30B-A3B-2507 (96.3\% to 94.7\%). Llama-3.1-70B achieved the highest accuracy for the shortest records (Q1), whereas Qwen3-30B-A3B-2507 achieved the highest accuracy for the longest records (Q4). In contrast, reasoning-oriented models in the R1-Distill family showed greater sensitivity to longer inputs, with R1-Distill-Llama-70B declining from 96.9\% to 91.8\%, R1-Distill-Qwen-32B from 94.7\% to 91.3\%, and R1-Distill-Qwen-7B from 61.3\% to 40.6\%. The Llama-Poro-2 variants showed pronounced deterioration on longer records, where EHR lengths exceeded their supported input length, with accuracy declining from 92.6\% to 6.2\% for Llama-Poro-2-70B and from 74.5\% to 0.9\% for Llama-Poro-2-8B. Because the quartiles differed not only in input length but also in patient cases and question distribution, these results reflect combined effects of record length and case composition rather than a purely causal effect of length alone.

\subsection*{Memory requirements and inference latency}
GPU memory consumption and inference latency varied substantially across models under bfloat16 precision (Table~\ref{tab:main_results}, Supplementary Fig.~1). Large models required peak memory allocations exceeding 130~GB, whereas smaller models operated below approximately 25~GB. Two models exceeded the available 160~GB memory limit and could not be further evaluated. Latency ranged from sub-second responses for smaller models to more than 150~s for the slowest configurations. Notably, the reasoning models showed markedly longer response times than generalist models of comparable size. DeepSeek-R1-Distill-Qwen-32B required up to 91~s per query and DeepSeek-R1-Distill-Llama-70B up to 153~s. Similarly sized generalist models typically responded within 20~s. The Finnish-language Llama-Poro-2-70B model also exhibited prolonged latency. In contrast, Qwen3-30B-A3B-2507 produced responses in under 10~seconds even for long inputs.

\begin{table*}[ht]
\centering
\footnotesize
\caption{\textbf{Quantization effects on clinical contextual question answering accuracy and memory usage.}
Accuracy is reported for the V1 formulation at three numerical precisions (bf16, 4-bit, and 8-bit). Peak 4-bit VRAM is reported as the minimum and maximum per-query values across the dataset. The best results per column are shown in bold. Models marked with $^{\dagger}$ were evaluated on a reduced subset ($N=955$) due to context window constraints. All other models were evaluated on the full dataset ($N=1{,}664$).}
\label{tab:quantization}
\begin{tabular}{llccccc}
\toprule
\multirow{2}{*}{\textbf{Developer}} &
\multirow{2}{*}{\textbf{Model}} &
\multicolumn{3}{c}{\textbf{Accuracy (\%) $\uparrow$}} &
\multicolumn{2}{c}{\textbf{4-bit VRAM (GB) $\downarrow$}} \\
\cmidrule(lr){3-5}\cmidrule(lr){6-7}
 &  & \textbf{bf16} & \textbf{4-bit} & \textbf{8-bit} &
\textbf{Min} & \textbf{Max} \\
\midrule
\multirow{4}{*}{Google}
 & Gemma-3-4B-it~\cite{2025gemma3} & 72.6 & 66.9 & 71.8 & 3.4 & 11.6 \\
 & Gemma-3-27B-it~\cite{2025gemma3} & 92.7 & 92.5 & 92.6 & 16.7 & 36.0 \\
 & MedGemma-4B-it~\cite{2025medgemma} & 74.5 & 74.0 & 74.3 & 3.4 & 11.6 \\
 & MedGemma-27B-text-it~\cite{2025medgemma} & 92.1 & 93.0 & 92.2 & 16.5 & 35.8 \\
\addlinespace
\multirow{3}{*}{DeepSeek}
 & R1-Distill-Qwen-7B~\cite{2025deepseek-r1} & 50.4 & 46.3 & 49.8 & 5.6 & 10.2 \\
 & R1-Distill-Qwen-32B~\cite{2025deepseek-r1} & 92.7 & 92.4 & 92.6 & 18.8 & 29.6 \\
 & R1-Distill-Llama-70B~\cite{2025deepseek-r1} & 94.8 & 81.3 & 11.2 & 38.7 & 53.6 \\
\addlinespace
\multirow{3}{*}{Meta}
 & Llama-3.1-8B~\cite{2024llama3}  & 79.9 & 72.6 & 79.3 & 6.0 & 13.0 \\
 & Llama-3.1-70B~\cite{2024llama3}  & \textbf{95.3} & 92.1 & 65.1 & 38.6 & 53.6 \\
 & Llama-3.3-70B~\cite{2024llama3}  & 94.9 & 84.6 & 61.4 & 38.6 & 53.6 \\
\addlinespace
\multirow{2}{*}{Qwen}
 & Qwen3-4B-2507~\cite{yang2025qwen3} & 91.1 & 85.3 & 91.5 & \textbf{3.0} & \textbf{10.0} \\
 & Qwen3-30B-A3B-2507~\cite{yang2025qwen3} & 95.1 & \textbf{94.1} & \textbf{94.5} & 15.9 & 20.5 \\
\addlinespace
\multirow{2}{*}{LumiOpen}
 & Llama-Poro-2-8B~\cite{poro2_2025}$^{\dagger}$ & 70.2 & 68.0 & 46.6 & 5.8 & 10.2 \\
 & Llama-Poro-2-70B~\cite{poro2_2025}$^{\dagger}$ & 90.6 & 90.9 & 6.8 & 38.2 & 40.5 \\
\bottomrule
\end{tabular}
\end{table*}

\subsection*{Accuracy and memory effects of low-precision quantization}
Low-precision weight quantization substantially reduced GPU memory requirements while largely preserving accuracy for most models (Table~\ref{tab:quantization}, Supplementary Figs.~2 and 3). For example, Llama-3.1-70B accuracy decreased from 95.3\% (bf16) to 92.1\% at 4-bit precision, while the peak memory footprint dropped from 148.2~GB to 53.6~GB. Similarly, Qwen3-30B-A3B-2507 retained high accuracy, decreasing from 95.1\% in bf16 to 94.1\% in 4-bit precision, while memory usage decreased from 61.8~GB to 20.5~GB. In contrast, several models exhibited pronounced accuracy degradation under 8-bit quantization despite intermediate memory savings. Notably, the accuracy of Llama-3.1-70B fell to 65.1\% in 8-bit precision, and R1-Distill-Llama-70B to 11.2\% compared to 94.8\% in bf16. The magnitude of degradation varied between model architectures, indicating that the quantization effects depend strongly on the model design and the quantization scheme. These results support 4-bit quantization as a practical strategy for deploying large models under memory constraints, with only a modest accuracy trade-off in most architectures.

\subsection*{Effects of prompt conditioning}
Here, two prompt conditioning strategies were evaluated: one-shot prompting, which provides a single example case in the prompt, and context conditioning, which adds extended explanations of the clinical task (Supplementary Table~1). In general, both strategies produced only modest changes relative to baseline performance. One-shot prompting yielded variable effects across models, with accuracy increasing for some model architectures and decreasing for others. For example, Gemma-3-4B-it decreased from 72.6\% to 69.5\%, while Qwen3-4B-2507 increased from 91.1\% to 92.3\%. Context conditioning produced similarly small changes, generally within one to two percentage points. Several higher-performing models showed slight improvements or no changes, including Llama-3.1-70B (from 95.3\% to 95.7\%) and Qwen3-30B-A3B-2507 (remained 95.1\%). Context-limited models could not be evaluated because the additional prompt content exceeded their supported input window. In summary, prompt conditioning did not produce consistent performance gains across model families.

\begin{table*}[ht]
\centering
\footnotesize
\caption{\textbf{Calibration performance for clinical contextual question answering under multiple-choice selection.} Accuracy is reported for the V1 question formulation. Calibration metrics include expected calibration error (ECE), maximum calibration error (MCE), and bias. The best results per column are shown in bold. Models marked with $^{\dagger}$ were evaluated on a reduced subset ($N=955$) due to context window constraints. All other models were evaluated on the full dataset ($N=1{,}664$). DeepSeek R1 models were excluded because their architecture requires intermediate reasoning generation.}
\label{tab:calibration_metrics}
\begin{tabular}{llcccc}
\toprule
\textbf{Developer} &
\textbf{Model} &
\textbf{Acc. (\%) $\uparrow$} &
\textbf{ECE $\downarrow$} &
\textbf{MCE $\downarrow$} &
\textbf{Bias $\downarrow\left|\cdot\right|$} \\
\midrule
\multirow{4}{*}{Google}
 & Gemma-3-4B-it~\cite{2025gemma3} & 72.5 & 0.238 & 0.397 & 0.114 \\
 & Gemma-3-27B-it~\cite{2025gemma3} & 94.1 & 0.045 & \textbf{0.277} & 0.023 \\
 & MedGemma-4B-it~\cite{2025medgemma} & 73.9 & 0.156 & 0.539 & \textbf{-0.007} \\
 & MedGemma-27B-text-it~\cite{2025medgemma} & 93.7 & 0.193 & 0.707 & -0.193 \\
\addlinespace
\multirow{3}{*}{Meta}
 & Llama-3.1-8B~\cite{2024llama3}  & 80.3 & 0.186 & 0.624 & -0.186 \\
 & Llama-3.1-70B~\cite{2024llama3}  & \textbf{95.7} & 0.123 & 0.561 & -0.123 \\
 & Llama-3.3-70B~\cite{2024llama3}  & 95.0 & \textbf{0.039} & 0.643 & 0.038 \\
\addlinespace
\multirow{2}{*}{Qwen}
 & Qwen3-4B-2507~\cite{yang2025qwen3} & 91.2 & 0.059 & 0.435 & 0.059 \\
 & Qwen3-30B-A3B-2507~\cite{yang2025qwen3} & 95.1 & 0.083 & 0.478 & -0.066 \\
 \addlinespace
\multirow{2}{*}{LumiOpen}
 & Llama-Poro-2-8B~\cite{poro2_2025}$^{\dagger}$ & 67.1 & 0.149 & 0.311 & 0.055 \\
 & Llama-Poro-2-70B~\cite{poro2_2025}$^{\dagger}$ & 92.7 & 0.068 & 0.511 & 0.068 \\
\bottomrule
\end{tabular}
\end{table*}

\subsection*{Multiple-choice accuracy and calibration}
The accuracy in multiple-choice selection was similar to that observed with free-text generation (Table~\ref{tab:calibration_metrics}). Relative performance rankings were largely preserved, with the highest accuracy achieved by larger models. This evaluation setting enabled direct assessment of probabilistic calibration and revealed substantial variability in confidence calibration across model architectures. High accuracy did not consistently correspond to well-calibrated predictions. For example, Llama-3.1-70B achieved the highest accuracy (95.7\%) but showed a moderate calibration error (ECE 0.123), while Llama-3.3-70B exhibited slightly lower accuracy (95.0\%) but the lowest ECE value (0.039). Gemma-3-27B-it combined high accuracy with strong calibration (ECE 0.045, MCE 0.277), while MedGemma-27B-text-it demonstrated comparable accuracy but markedly poorer calibration (MCE 0.707). Smaller models generally exhibited larger calibration errors. The bias values indicated heterogeneous tendencies toward overconfidence or underconfidence, and several high-accuracy models showed mild underconfidence. Models evaluated on a reduced dataset due to context window constraints showed similar qualitative patterns but are not directly comparable. In general, restricting responses to predefined options revealed substantial differences in the reliability of model confidence estimates without materially affecting accuracy.

\subsection*{Clinical significance of model errors}
To assess the clinical significance of model errors, we conducted two complementary analyses. First, all V1 responses flagged as incorrect by the autograder (78 out of 1,664) were reviewed by a medical oncologist. Expert adjudication reclassified 15 responses as correct, 15 as having clinically insignificant errors, and 48 as having clinically significant errors, thus indicating that approximately 3\% of all V1 responses contained clinically significant errors. Second, we examined cases in which semantically equivalent V1 and V2 questions produced inconsistent responses, focusing on instances where one response was correct and the other incorrect. Of 38 pairs of this type, three were reclassified as correct after expert review, leaving 35 pairs. Within these, 16 pairs (0.96\% of all 1,664 V1 questions) included one correct response and one clinically significant error, and 19 pairs (1.14\%) included one correct response and one clinically insignificant error.


\section{Discussion}\label{Discussion}

In this study we have evaluated 14 locally deployable, instruction-tuned open-source large language models for Clinical Contextual Question Answering (CCQA) across complete longitudinal electronic health records in a Finnish oncology setting. Several long-context generalist models demonstrated high accuracy and strong agreement across semantically equivalent question formulations. This suggests that patient-specific clinical information can be accurately retrieved from heterogeneous Finnish EHR narratives when complete records are processed without retrieval-based preselection. All experiments were performed within a closed offline computing environment without external data transmission. By assessing accuracy, response consistency, probabilistic calibration, inference latency, GPU memory requirements, quantization behaviour, and clinician-adjudicated safety, this work provides a deployment-oriented characterization of model performance under realistic offline constraints.

Prior work on medical natural language processing has focused primarily on medical question-answering benchmarks or narrowly scoped extraction tasks~\cite{singhal2023large, thirunavukarasu2023large}. In this study, we address patient-specific reasoning in complete longitudinal records in an under-represented clinical language setting, specifically EHRs in Finnish, where clinically relevant information may be distributed across years of documentation. Rather than segmenting records using retrieval pipelines~\cite{lewis2020retrieval}, we examined retrieval-free ingestion of complete patient histories, enabling assessment of contextual integration across temporally separated clinical events. This design reflects clinical scenarios in which relevant information may be dispersed and not easily identifiable through local similarity-based retrieval alone. The observed performance suggests that long-context instruction-tuned models can support privacy-preserving clinical information retrieval within institutionally governed infrastructure~\cite{ghassemi2020review, world2024ethics}.

For large language models, performance generally improves with scale~\cite{kaplan2020scaling, hoffmann2022training}. However, we found that parameter count alone did not determine clinical task performance. A mid-sized 30B-parameter model achieved accuracy comparable to the leading 70B models, while requiring substantially less memory and lower latency, suggesting that architectural efficiency and instruction tuning influence downstream clinical utility. Furthermore, reasoning-oriented LLMs~\cite{wei2022chain, 2025deepseek-r1} reached similar accuracy but incurred longer inference times due to intermediate reasoning steps, without consistent gains in final answer accuracy. In time-constrained clinical workflows, such latency may limit usability~\cite{ghassemi2020review}. Thus, model selection in healthcare settings should jointly consider predictive performance, computational efficiency, and workflow compatibility.

Agreement across semantically equivalent question formulations was high among the top-performing systems, with several models achieving consistency rates between 97\% and 98\%. Such stability is desirable in clinical settings, where queries can be phrased in multiple ways~\cite{ghassemi2020review}. Nevertheless, our clinical safety analysis revealed that paraphrasing can produce divergent responses in some cases, with some formulations yielding correct answers while others lead to clinically significant errors. The robustness to linguistic variation, therefore, remains only partially resolved and should be considered when integrating LLMs into clinical workflows. Because model outputs may depend on the precise wording of an input, evaluation frameworks for clinical language models should account for potential sensitivity to question formulation.

For individual clinical questions, the best-performing models achieved high accuracy, with question-specific performance ranging from 87\% to 100\% (Supplementary Tables~2 and~3). Questions involving categorical attributes typically documented in pathology reports or treatment summaries, such as tumour laterality or invasiveness, histological subtype, and receptor status, were answered with consistently high accuracy. In contrast, lower performance was observed in tasks requiring temporal reasoning over treatment timelines or interpretation of quantitative values in electronic health records. For example, determining whether a patient received neoadjuvant therapy (i.e., treatment administered before surgery), requires integrating temporally separated information. These patterns suggest that long-context language models perform most reliably when clinical variables are explicitly documented, whereas tasks requiring contextual interpretation across longitudinal documentation remain more challenging.

Models also differed in their deployment characteristics. The maximum usage of GPU memory for long EHR input ranged from approximately 15 GB to exceeding the available 160 GB memory limit. The inference latency varied similarly widely, from sub-second responses to several minutes per query. In clinical encounters in which multiple queries may be issued, response times measured in minutes may be incompatible with routine clinical workflows. These findings highlight that the selection of models for clinical deployment should consider not only predictive performance but also GPU memory requirements and inference latency. In particular, the Qwen3-30B model achieved accuracy comparable to the best 70B-parameter models while requiring substantially lower GPU memory and inference time.

We found that performance varied with increasing EHR length, indicating heterogeneous robustness to extended clinical documentation. Several long-context models maintained relatively high accuracy across record-length quartiles with only modest declines, whereas others showed greater sensitivity to longer inputs. These findings suggest that nominal support for long context windows does not guarantee stable performance on long clinical records. Because the record-length quartiles comprised different patient cases, the observed differences likely reflect combined effects of input length and case composition rather than a purely causal effect of record length. Overall, these results indicate that strong performance on short-text tasks does not necessarily generalize to longer electronic health records, even when models nominally support context lengths exceeding the record length.

Quantization reduced memory requirements while largely preserving predictive performance. In particular, 4-bit quantization enabled substantial reductions in GPU memory usage with only modest declines in accuracy across several architectures, improving feasibility for deployment on lower-cost hardware. However, the effects of quantization varied substantially between models. Some architectures maintained accuracy comparable to bf16, whereas others exhibited pronounced degradation. For example, the accuracy of Llama-3.1-70B decreased from 95.3\% to 65.1\%, and R1-Distill-Llama-70B decreased from 94.8\% to 11.2\% under 8-bit quantization. These findings indicate that quantization can substantially improve computational efficiency but does not guarantee preserved performance. Quantized configurations should therefore be empirically validated prior to clinical deployment.

We also found that the calibration analysis revealed variability in the alignment between model confidence and empirical accuracy. High predictive accuracy did not consistently correspond to well-calibrated probability estimates, and some architectures demonstrated more reliable calibration than others~\cite{nakkiran2025trained}. In clinical settings, confidence estimates from well-calibrated models can help determine when the output is likely reliable and when clinician verification is required. Therefore, calibration metrics provide complementary information to accuracy when selecting and validating models for clinical deployment.

Neither language specialization nor medical domain specialization conferred an advantage in our evaluation. Models adapted for medical text or specialized for Finnish language did not outperform generalist models at comparable parameter scales when answering patient-specific clinical questions from EHRs. These results suggest that broad multilingual pretraining may already provide sufficient representational capacity to interpret heterogeneous clinical documentation written in Finnish. These findings highlight the importance of evaluating models directly in the target clinical tasks rather than assuming advantages from domain or language specialization.

Clinical safety evaluation provided additional insight into the potential risks associated with model-generated responses. The clinical safety analysis was performed on the highest-performing model (Llama-3.1-70B). Although the overall accuracy was 95.3\%, expert review showed that 2.8\% of all responses contained clinically significant errors. In addition, semantically equivalent questions occasionally produced divergent responses, with one formulation yielding a correct answer and another a clinically significant error (0.96\% of responses). These findings provide empirical evidence that high overall accuracy does not capture the clinical severity of errors in LLM outputs, and that clinically significant errors may still occur.

This study has several limitations. The evaluation was conducted using a dataset of 183 patient records from a single Finnish hospital region and focused on one oncology domain, which may limit the generalizability of the findings to other languages, clinical domains, healthcare settings or documentation practices. The clinical records were primarily in Finnish and contained embedded medical terminology in English and Latin typical of routine documentation, and therefore the generalizability in other linguistic or healthcare settings remains uncertain. Raw clinical texts cannot be publicly released due to privacy constraints, which limits full external reproducibility.

The task focused on predefined clinical variables with restricted answer options and may not capture the full scope of open-ended clinical queries or diagnostic reasoning. Although the models generated free-text responses, predefined answer options were embedded within the prompt to encourage standardized terminology and enable reliable automated classification aligned with the annotated label space. This design may have influenced the formulation of the response and increased the observed consistency relative to the answer of fully unconfined questions. The study was retrospective and did not evaluate prospective workflow outcomes such as time savings, reduction of cognitive load, or impact on clinical decision-making. Annotation relied on a single expert with repeat labelling to assess intra-annotator reliability, and residual labelling bias cannot be excluded. Deterministic decoding was used to ensure reproducible and auditable behaviour consistent with clinical validation requirements, which means that identical inputs always produced the same output without stochastic variation. Deployments using stochastic decoding strategies may exhibit greater response variability and would require independent evaluation.

In summary, several locally deployable open-source LLMs were able to accurately retrieve clinically relevant patient-specific information from longitudinal Finnish EHRs in response to natural-language clinical questions without requiring external data transmission. However, predictive performance alone does not determine readiness for clinical deployment. The presence of clinically significant errors indicates that such systems should be deployed as clinician-support tools rather than autonomous decision-makers, consistent with prior work showing that LLMs may produce clinically incorrect outputs and hallucinations despite strong overall performance~\cite{asgari2025framework,roustan2025clinicians,shool2025systematic}. Clinical deployment further depends on factors including consistency across semantically equivalent queries, calibration, latency, memory requirements, and quantization stability. Prospective multi-site validation and workflow-integrated evaluation in real-world clinical settings are essential next steps. By jointly analysing predictive performance, computational feasibility, and clinician-adjudicated safety within a privacy-preserving offline framework, this study provides practical guidance for integrating large language models into routine clinical information retrieval workflows.

\section{Methods}
\label{methods}

We conducted a retrospective observational study to evaluate automated extraction of clinically relevant breast cancer information from electronic patient records using instruction-tuned large language models. The study utilized routinely collected clinical documentation generated during diagnostic evaluation, treatment, and follow-up care in Finnish hospital systems. Eligible records corresponded to patients with undergoing breast cancer screening or treatment. All analyses were performed on pseudonymised data within a secure, access-controlled offline computing environment compliant with institutional data protection policies and the General Data Protection Regulation (GDPR)~\cite{eu2016gdpr}. The study did not influence patient management and involved no direct patient contact. The study was approved and the need for patients' informed consent was waived by the Institutional Review Board of the Wellbeing Services County of Pirkanmaa (permission code: R24221).

\subsection{Dataset and annotation}

We assembled a retrospective dataset of 183 pseudonymised patient records from the Wellbeing Services County of Pirkanmaa, Finland. The cohort included individuals with current or prior breast cancer, suspected disease, or screening encounters without malignant findings. The cohort included patients with current and previous breast cancers, and records also contained information on other medical conditions and treatments. Data were extracted by the data-providing authority in accordance with data-minimization principles to include only clinical information relevant to breast cancer diagnosis, treatment, surveillance, and related investigations. The resulting dataset was delivered in pseudonymised form and analysed within a secure offline research environment. A detailed description of the data structure is provided in Supplementary Section~A.

Patient records were represented as a standardized table of 26 predefined clinical fields (key–value pairs), where individual field values ranged from coded entries to extended narrative text spanning multiple pages. These included pseudonymised patient identifiers, temporal information on clinical events, coded clinical entries, and free-text narratives describing imaging, pathology, surgical procedures, systemic therapies, medications, and information of follow-ups. Documentation was primarily in Finnish, with occasional English and Latin medical terminology reflecting routine clinical practice. No translation or language normalization was applied. Patient-specific observation periods ranged from 2 to 62 years, with a median of 14 years. Record length ranged from 408 to 5,508 words, with a median of 1,792 words. This corresponded to 3,828 to 51,007 characters, with a median of 17,365 characters including spaces.

Each record was annotated using up to 21 predefined clinical question–answer pairs addressing breast cancer history, tumour characteristics, receptor status, laterality, nodal involvement, metastatic disease, and primary treatments. Responses were restricted to categorical labels or numeric values, with an “Unknown” category assigned when relevant information was absent or inconclusive. Questions not applicable to a given patient were omitted. For example, tumour-specific or treatment-related questions were not applied to patients without confirmed invasive breast cancer. Annotations were restricted to clinical events occurring in or after 2015 to establish a consistent temporal reference window across patients with potentially multiple cancer episodes. An exception was CQ1 (year of first breast cancer diagnosis), which required identification of the earliest documented breast cancer regardless of the year of the diagnosis. Clinical documentation predating 2015 was retained because accurate determination of post-2015 clinical variables often depended on prior diagnostic history, treatments, and longitudinal disease trajectories. The same temporal constraints were incorporated into the prompts provided to the models, ensuring alignment between annotation criteria and model instructions. Detailed definitions of all questions and response categories are provided in Supplementary Section~B.

Annotations were performed by a medical oncologist with 6 years of specialization in breast cancer. To assess intra-annotator reliability, the annotator repeated the labelling process after a two-month interval while blinded to the initial annotations. Of the 1,927 question–answer instances annotated in both rounds, 1,685 (87.4\%) were identical. Instances with discordant labels were excluded from further analysis.

From these high-confidence annotations, 21 instances (one per clinical question) were reserved for framework development and construction of a single one-shot demonstration example included in the prompt. These instances were not part of the evaluation dataset. The final evaluation dataset therefore comprised 1,664 question–answer pairs.

All records were pseudonymised by the data-providing authority prior to transfer. Direct identifiers were removed and replaced with study-specific pseudo-identifiers, and the linkage key enabling re-identification was retained exclusively by the data controller and was not accessible to the investigators. The dataset therefore constituted pseudonymised personal data under the General Data Protection Regulation (GDPR). Data processing and analysis were conducted within a secure, access-controlled offline environment compliant with GDPR requirements. Due to privacy constraints, neither the original identifiable records nor the pseudonymised clinical texts used in this study can be publicly released. Synthetic examples that preserve the structure and key characteristics of the dataset are provided in Supplementary Section~A.

\subsection{Models evaluated}

Sixteen instruction-tuned large language models were evaluated, spanning general-purpose (Gemma-3, Llama-3.x, Qwen3), medically adapted (MedGemma), Finnish language-specialised (Poro-2), and reasoning-oriented architectures (DeepSeek R1, Llama-4-Scout-17B, GPT-OSS-20B) (Table~\ref{tab:main_results}). Two models (Llama-4-Scout-17B and GPT-OSS-20B) exceeded the 160,GB GPU memory limit and were therefore excluded from the main experiments. Detailed memory profiling is provided in Supplementary Section~I.

Most models supported input context lengths exceeding those of the patient records analysed in this study. The Finnish language-specialised Poro-2 models were the exception, with a maximum input length of 8,192 tokens, and were retained to represent language-specialised architectures despite this constraint.

All models, weights, and tokenizers were obtained from official Hugging Face repositories without modification~\cite{wolf2020transformers}. Repository identifiers and exact commit hashes used for inference are provided in Supplementary Section~G to ensure full reproducibility.

\subsection{Experimental Setup}

Experiments were conducted in a secure offline computing environment using an AMD EPYC 75F3 32-core CPU, 256 GB RAM, and two NVIDIA A100 GPUs with 80 GB VRAM each (160~GB total GPU memory). Models were executed using the Hugging Face Transformers framework~\cite{wolf2020transformers}, with low-bit quantization implemented via bitsandbytes~\cite{dettmers2022llmint8} where supported.

Prior to the main experiments, GPU memory profiling was performed to determine which models could be deployed within the available hardware constraints. Profiling was conducted in an open high-performance computing environment using higher-capacity hardware to characterize memory requirements across a wide range of input lengths. Synthetic input sequences of controlled length were used to measure peak memory consumption during a single forward decoding step, with sequence lengths ranging from minimal input to 25{,}120 tokens in increments of 512 tokens. Model inclusion in the main experiments was determined by feasibility within the 160~GB deployment budget. Profiling was performed without clinical data, and each model was evaluated in an isolated process to ensure accurate measurement. Additional details and full results are provided in Supplementary Section~I.

For each evaluation instance, the complete patient record was provided to the model without prior filtering or retrieval. Records consisted of all predefined clinical fields formatted as a single text document by concatenating column names and their contents as key–value pairs separated by blank lines. Individual field values ranged from short coded entries to extended narrative text. Data were presented exactly as stored in the source records. The full prompt template is provided in Supplementary Section~C.

Models were evaluated using a question format with predefined answer options included in the prompt to encourage use of the target terminology and to reduce scoring errors arising from lexical variation, synonyms, or language differences. Answer options were ordered alphabetically to avoid positional bias or unintended information leakage.

Inference was performed in bfloat16 precision for all models, except in dedicated analyses of low-precision deployment. Additional experiments using 8-bit and 4-bit quantization were conducted only for models compatible with the bitsandbytes library. To obtain deterministic outputs, stochastic decoding was disabled by setting temperature to zero and disabling sampling and beam search. Models were instructed to produce only the final answer without explanation, resulting in short responses typically consisting of a single word or numeric value. Maximum output length was limited to 25 tokens for standard instruction-tuned models to allow minor verbosity without truncating valid responses. Reasoning-oriented models that generate intermediate reasoning traces prior to the final response were allowed up to 1{,}000 tokens to ensure completion of the reasoning process, although only the final answer was used for evaluation.

All models were evaluated using their native tokenizers and instruction-tuned checkpoints. Identical prompts, patient records, and evaluation scripts were used across all experiments to ensure consistency and reproducibility.

\subsection{Prompting Strategies}

All experiments employed structured prompts comprising task instructions and input data, including the patient EHR record, clinical questions, and predefined answer options. Prompts were constructed using each model’s native chat template to ensure consistent formatting across models. Task instructions specified an expert mammography radiologist role and directed the model to base responses solely on the patient record, select the single most accurate option from predefined choices, and return only the final answer without explanation. Predefined categorical or numeric answer options were included to guide responses toward standardized terminology, with an \textit{Unknown} option used when relevant information was absent or inconclusive. Temporal constraints consistent with the annotation protocol were specified in the task instructions. These instructed the model to restrict responses to events occurring in or after 2015, except for the first breast cancer diagnosis, which required identifying the earliest documented date.

Patient records were provided in full without filtering or retrieval and were represented as a single text by concatenating all available data fields. Prompts were applied consistently across both free-text generation and multiple-choice tasks to maintain comparability (Supplementary Section~C). 

To examine the influence of prompt design on model performance and robustness, several specialized prompting strategies were evaluated:
\begin{itemize}
\item \textbf{Task-specific context:} Medically relevant background information was prepended to provide additional guidance for clinically nuanced questions (e.g., definitions of invasive cancer or tumour laterality).
\item \textbf{One-shot prompting:} A single in-context example consisting of a patient record, question, answer options, and correct answer was included, drawn from data not used in the evaluation set.
\item \textbf{Consistency testing:} Target questions were replaced with semantically equivalent paraphrases to assess robustness to variations in wording while preserving clinical meaning.
\end{itemize}

This framework enabled controlled comparison of prompt configurations while maintaining identical task requirements.

\subsection{Evaluation and Metrics}
\label{evaluation_and_metrics}

Models were evaluated under two configurations: free-text generation and multiple-choice selection. In the free-text generation setting, models produced open-ended responses, reflecting standard generative LLM behaviour. Decoding was deterministic (temperature = 0, sampling disabled), ensuring identical outputs for identical inputs and supporting reproducible and predictable behaviour, consistent with guidance of AI systems for health~\cite{world2024ethics}. Generated text was graded according to the procedure described below.

In the multiple-choice selection setting, models selected from a set of $K$ predefined answer options $\{O_1, \dots, O_K\}$. Prompts and patient records were identical across both settings. For each option $i$, we computed a length-normalized score $s_i$ as the average log-probability of its constituent tokens given the prompt. To enable calibration analysis, these scores were converted into a probability distribution using the softmax function:

\begin{equation}
P(O_i) = \frac{\exp(s_i)}{\sum_{j=1}^{K} \exp(s_j)}
\end{equation}

The option with the highest probability was selected as the model’s response, and its corresponding probability \(\max\limits_{i} P(O_i)\) was recorded as the model’s confidence for that prediction.

\subsubsection*{Rule-based substring baseline}
For comparison, we included a deterministic rule-based method based on case-insensitive substring matching (Rule-based (substring)). Each answer option was searched directly within the patient record text, and the option with the highest match frequency was selected as the prediction. This approach relies solely on surface-level text matching and does not incorporate contextual understanding, temporal reasoning, or cross-sentence inference.

\subsubsection*{Answer Matching and Metrics}
Performance was assessed using instance-level binary grading. Each patient--question pair constituted one evaluation instance, and model outputs were scored as either correct or incorrect. Model outputs were graded in a case-insensitive manner against the reference answers. A response was graded as correct if the target answer option could be identified in the model output and no incorrect alternatives were present. Responses containing explanatory text or additional content were accepted provided that this condition was satisfied. Otherwise, the response was graded as incorrect.

All question--answer pairs were designed to yield either a numeric value such as year or tumor size or a categorical label from a predefined set such as left, right, or bilateral. These answer formats were explicitly specified in the prompt to guide models toward standardized outputs suitable for automated grading. Hierarchical labels such as N1 and N1a were matched using specificity-aware rules to prevent misclassification due to partial matches.

Overall accuracy was computed as the proportion of correctly graded instances:
\begin{equation}
\text{Accuracy} = \frac{N_{\text{correct}}}{N_{\text{total}}} .
\end{equation}

Results are reported as overall accuracy across the dataset and as per-question accuracies for the 21 clinical variables (Supplementary Section~E). To quantify statistical uncertainty, 95\% confidence intervals for accuracy were computed using the exact Clopper–Pearson method for binomial proportions (Supplementary Section~G).

Statistical differences in accuracy between models were evaluated using paired McNemar tests applied to binary correctness outcomes (correct vs. incorrect) for each question--answer pair under the Version~1 (V1) question formulation. Comparisons were performed on matched predictions for the same instances across models. The test used the exact binomial formulation based on the number of discordant instances ($n_{10}$ and $n_{01}$) between two models. To control for multiple comparisons across model pairs, $P$ values were adjusted using the Holm correction. Detailed contingency counts and adjusted $P$ values are reported in Supplementary Section~G.

\subsubsection*{Consistency Analysis}
To assess robustness to linguistic variation in query phrasing, each clinical question was expressed in two semantically equivalent formulations referring to the same underlying clinical variable, denoted Version~1 (V1) and Version~2 (V2). The two versions differed only in wording while preserving identical clinical meaning and target labels. Both V1 and V2 were evaluated independently for each applicable patient record, and performance was computed separately for each version.

Consistency was defined as agreement between the answers extracted from the model outputs for V1 and V2, with answer assignment performed as described in \ref{evaluation_and_metrics}. An instance was classified as consistent if both versions yielded the same answer and inconsistent otherwise. Overall consistency was reported as the proportion of instances with identical responses across versions, irrespective of correctness.

\subsubsection*{Calibration Analysis}
Calibration was evaluated in the multiple-choice selection setting by assessing the agreement between predicted confidence and observed accuracy. Confidence for each prediction was defined as the maximum softmax probability over the predefined answer options, \(\max\limits_{i} P(O_i)\), computed from length-normalized token log-probabilities.

The primary metric was the \textbf{Expected Calibration Error (ECE)}~\cite{guo2017calibration}, which quantifies the weighted average discrepancy between confidence and accuracy across $M=10$ equally spaced confidence bins over the interval $[0,1]$:
\begin{equation}
\text{ECE} = \sum_{m=1}^{M} \frac{|B_m|}{N} \left| \text{acc}(B_m) - \text{conf}(B_m) \right| ,
\end{equation}
where $N$ is the total number of instances and $B_m$ denotes the subset of predictions in bin $m$.

We also report the \textbf{Maximum Calibration Error (MCE)}, defined as the largest calibration gap across bins, and \textbf{Calibration Bias}, defined as the difference between mean predicted confidence and mean accuracy, where positive values indicate overconfidence, negative values indicate underconfidence, and values closer to zero indicate better calibration.

\subsubsection*{Clinical Safety Evaluation}

To evaluate potential clinical risk associated with LLM-generated responses, we conducted a blinded expert review of outputs from the highest-performing model in the free-text generation setting (Llama-3.1-70B, Version~1 accuracy 95.3\%). The assessment comprised two complementary analyses. First, all V1 responses flagged as incorrect by the automated grader (78 of 1,664) were reviewed to determine the clinical significance of errors. Second, all patient-specific question–answer pairs in which semantically equivalent V1 and V2 questions produced inconsistent responses were included, focusing on instances in which one response was correct and the other incorrect (38 of 1,664 V1+V2 pairs).

All cases were presented in random order prior to assessment. A medical oncologist with 6 years of subspecialty experience in breast cancer, who had performed the original annotations, reviewed each case while blinded to patient identifiers and to the reference annotation labels. Only the model’s final answer text generated under free-text generation was evaluated. The reviewer had access to the patient record and the corresponding clinical question for each case.

Each response was classified into one of three mutually exclusive outcome categories:

\begin{itemize}
\item \textit{Correct:} The response accurately reflects the clinical information in the record without clinically relevant errors.
\item \textit{Clinically significant error:} An error with potential to influence diagnostic reasoning, treatment decisions, or patient management.
\item \textit{Clinically insignificant error:} An inaccuracy unlikely to affect clinical judgment or patient care.
\end{itemize}

\subsection{Ethical Considerations and Clinical Safety}
Ethical approval was obtained from the Institutional Review Board of the Wellbeing Services County of Pirkanmaa (permission code R24221), with patient consent waived in accordance with Finnish law and institutional policy. All patient records were fully pseudonymised at the hospital before transfer to a secure, offline environment, in compliance with GDPR~\citep{eu2016gdpr}. Models were assessed solely for research purposes. All illustrative examples are synthetic to maintain patient privacy.

\backmatter

\bmhead{Supplementary information}
Supplementary material provides additional methodological detail and complete results that support the main text. It includes the full structure and definitions of the structured EHR input table used to construct model prompts, with synthetic examples illustrating typical content in Section A. It also contains the complete Clinical Contextual Question Answering (CCQA) question set from CQ1 to CQ21, including wording variants, answer options, and curated clinical definitions used for expert annotation and optional prompt conditioning in Section B. Per question performance is reported by clinical question type for the primary wording variant V1 across all evaluated models and a deterministic keyword matching baseline in Supplementary Tables 1 and 2. The exact prompt template and serialization procedure used for consistent evaluation across models are provided in Section D, including the optional one-shot and task-specific context variants. Reproducibility metadata for every model are listed with Hugging Face repository identifiers and commit hashes in Table 3. GPU memory profiling experiments that quantify memory usage as a function of sequence length under the 160~GB budget are described in Section F and shown in Figures 1 to 3.

\bmhead{Acknowledgements}
This work was supported by Business Finland under the project \textit{Medical AI and Immersion} (2024–2026, decision number 10912/31/2022), Tampere University Hospital (\textit{State Research Funding for university-level health research / OOO project} (2022–2024)), the \textit{Cancer Foundation Finland} (2024–2025), and the \textit{Finnish Medical Foundation} (2023–2026). The funders had no role in study design, data collection and analysis, decision to publish, or preparation of the manuscript.

\bmhead{Author contributions}
M.S. contributed to conceptualization, study design, methodology development, software implementation, formal analysis, and writing of the main manuscript. N.H. contributed to study design, data collection, data curation, and figure preparation. O.A. provided clinical supervision, performed data annotation, and conducted the clinical safety evaluation of model outputs, and contributed to manuscript editing. J.S. and K.K. contributed to conceptualization and manuscript editing. All authors reviewed the manuscript.

\bmhead{Competing interests}
The authors declare no conflicts of interest.

\bmhead{Data availability}
The clinical datasets analysed during the current study are not publicly available due to data protection regulations (GDPR) and institutional policies of the Wellbeing Services County of Pirkanmaa governing patient data access. Synthetic data samples and record schemas are provided in the supplementary information.

\bmhead{Code availability}
The source code for the CCQA pipeline and evaluation scripts is available on GitLab at: version.aalto.fi/gitlab/saukkom3/mammo-llm.

\begin{appendices}

\clearpage
\section*{Supplementary Information}

\noindent \textbf{Clinical Contextual Question Answering with Locally Deployable Large Language Models for Longitudinal Electronic Health Records.} This file contains supplementary tables, figures, and methods supporting the main manuscript.

\subsection*{Contents}
\begin{enumerate}
    \item[A.] Structure of the Input Data
    \item[B.] Question Set
    \item[C.] Prompt Template and Construction
    \item[D.] Prompt Conditioning Experiments
    \item[E.] Per-question Accuracy by Clinical Question
    \item[F.] Performance as a Function of EHR Length
    \item[G.] Statistical Analysis
    \item[H.] Model Reproducibility
    \item[I.] GPU Memory Profiling
\end{enumerate}

\section*{A. Structure of the Input Data}
\label{supp:input_structure}
This section describes the structure of the input data used in the retrieval of answers to clinical questions. The data is formatted as a structured table, where each column contains either categorical metadata or free-form clinical text. All columns and their definitions were explicitly included in the input prompt to ensure that language models had full access to relevant contextual information.

These structured patient records served as the basis for manual annotation and were necessary to answer all defined clinical questions. The same column schema was used across all experiments to ensure consistency and comparability between expert-provided labels and model predictions.

Below, we document each column and provide synthetic examples to illustrate typical content. Real patient data cannot be published due to privacy regulations. All examples are presented in English for readability; however, the actual model inputs used in our experiments were in Finnish, English, or Latin, without translation.

\begin{enumerate}[label=\arabic*.]
    \item \textbf{Column:} maligni\_patlaus \\
          \textbf{Description:}  maligni\_patlaus column contains pathologists' reports regarding tissue samples, primarily from patients with breast cancer, but also includes results of samples from other anatomic locations. The entries detail findings related to malignancy, including carcinoma type, presence or absence of malignancy in specified locations (e.g., in breast or lymph nodes), histological grades (e.g., I-III), receptor status (e.g., estrogen receptor (ER), progesterone receptor (PgR), Human Epidermal Growth Factor Receptor 2 (HER2) expression status, presence of metastases, and specific diagnostic codes or classifications. This information represents a mix of diagnostic results, histological descriptions, and assessment of malignancy status. \\
          \textbf{Example:} (2020-01-01: PAD Axillary lymph node; (right): No evidence of malignancy; (0/5)), (2020-01-01: PAD Female breast, right: Ductal carcinoma (gradus 1) Axillary lymph node; (right): No evidence of malignancy; (0/5))
          
    \item \textbf{Column:} maligni\_pat\_kertomus \\
          \textbf{Description:} maligni\_pat\_kertomus column contains pathology reports, primarily concerning breast tissue, lymph nodes, and sometimes other organs. It details findings from histological examinations of biopsies, fine-needle aspirations, or surgical specimens, including descriptions of tumor type, presence or absence of malignancy in specified locations (e.g., in breast or lymph nodes), histological grades (e.g., I-III), receptor status (e.g., estrogen receptor (ER), progesterone receptor (PgR), Human Epidermal Growth Factor Receptor 2 (HER2) expression status, presence of metastases, and specific diagnostic codes or classifications. \\
          \textbf{Example:} (2020-01-01: PAD Axillary lymph node; (right): No evidence of malignancy; (0/5)), (2020-01-01: PAD Female breast, right: Ductal carcinoma (gradus 1) Axillary lymph node; (right): No evidence of malignancy; (0/5))
          
    \item \textbf{Column:} rintasyopa\_patlaus \\
          \textbf{Description:} rintasyopa\_patlaus column contains detailed histopathological reports of breast cancer diagnoses, often following surgical resection or biopsy. It includes information on tumor type, presence or absence of malignancy in specified locations (e.g., in breast or lymph nodes), histological grades (e.g., I-III), receptor status (e.g., estrogen receptor (ER), progesterone receptor (PgR), Human Epidermal Growth Factor Receptor 2 (HER2) expression status, presence of metastases, and specific diagnostic codes or classifications.\\
          \textbf{Example:} (2020-01-01: el Doctor Doctor/ (2020-01-01: PAD Axillary lymph node; (right): No evidence of malignancy; (0/5)), (2020-01-01: PAD Female breast, right: Ductal carcinoma (gradus 1) Axillary lymph node; (right): No evidence of malignancy; (0/5))
          
    \item \textbf{Column:} rintasyopa\_pat\_kertomus \\
          \textbf{Description:} rintasyopa\_pat\_kertomus column contains histopathological reports from breast cancer biopsies and resections, detailing surgical procedures and tumor characteristics. It includes findings such as tumor type, presence or absence of malignancy in specified locations (e.g., in breast or lymph nodes), histological grades (e.g., I-III), receptor status (e.g., estrogen receptor (ER), progesterone receptor (PgR), Human Epidermal Growth Factor Receptor 2 (HER2) expression status, presence of metastases, and specific diagnostic codes or classifications.\\
          \textbf{Example:} (2020-01-01 12:12:12 PAT 1.1.2020 1234 Br-PAD-2 1.1.2020 /Doctor Doctor / Pyyntö Rinnan histologinen tutkimus Ottopvm: 1.1.2020 50-vuotias nainen, jolla on molemmilla puolilla todettu rintasyöpä.)
          
    \item \textbf{Column:} rintasyopa\_kert\_muut \\
          \textbf{Description:} rintasyopa\_kert\_muut column contains free-text clinical notes documenting patient encounters related to breast cancer care. The records detail diagnoses, treatment history including surgeries and hormonal therapies, consultations, follow-up visits, and discussions regarding risks and management strategies. The entries are time-stamped, providing a longitudinal view of the patient's journey, and also include information about family history and other comorbidities. This data represents a rich source of information regarding the comprehensive clinical management of breast cancer patients.\\
          \textbf{Example:} (2020-01-01 12:12:12: Erotusdiagnoosina tulee kyseeseen rintakarsinooma,), (2020-01-01 12:12:12 Suvussa syöpään liittyviä mutaatioita: äidille tehtiin rinnanpoisto 50-vuotiaana.)
          
    \item \textbf{Column:} kokopoistot\_patlaus \\
          \textbf{Description:} kokopoistot\_patlaus column contains detailed histological and pathological findings from breast tissue samples. It includes tumor ttumor type, presence or absence of malignancy in specified locations (e.g., in breast or lymph nodes), histological grades (e.g., I-III), receptor status (e.g., estrogen receptor (ER), progesterone receptor (PgR), Human Epidermal Growth Factor Receptor 2 (HER2) expression status, presence of metastases, and specific diagnostic codes or classifications. \\
          \textbf{Example:} (2020-01-01: 1.1.2020 / Doctor Doctor/ pyyntö RINNAN HISTOLOGINEN TUTKIMUS, LAAJA LEIKKAUSPREPARAATTI Ottopvm: 1.1.2020 12.12. tehty onkoplastinen resektio oikeaan rintaan siten, että poistettu koko rinnan alamediaalineljännes nännin kanssa karsinooman vuoksi.)
          
    \item \textbf{Column:} kokopoistot\_pat\_kertomus \\
          \textbf{Description:} kokopoistot\_pat\_kertomus column contains detailed pathology reports, primarily focusing on breast cancer diagnoses and treatments. It includes information on tumor type, presence or absence of malignancy in specified locations (e.g., in breast or lymph nodes), histological grades (e.g., I-III), receptor status (e.g., estrogen receptor (ER), progesterone receptor (PgR), Human Epidermal Growth Factor Receptor 2 (HER2) expression status, presence of metastases, specific diagnostic codes or classifications, and specific treatment recommendations or plans, such as the type of surgery (mastectomy, breast-conserving surgery), adjuvant therapies (radiotherapy, chemotherapy, hormonal therapy), and follow-up care. It also includes details of patient history related to cancer, such as family history, previous treatments, and relevant medical conditions. \\
          \textbf{Example:} (2020-01-01 12:12:12 Hoidettu resektiolla ja vartijaimusolmuketutkimuksella 1.1. vasemmanpuoleinen alalateraalinen pT1a N0 gradus I duktaalinen invasiivinen rintasyöpä.)
          
    \item \textbf{Column:} osapoistot\_patlaus \\
          \textbf{Description:} osapoistot\_patlaus column contains detailed pathology reports, primarily focusing on breast cancer diagnoses and treatments. It includes information on tumor type, presence or absence of malignancy in specified locations (e.g., in breast or lymph nodes), histological grades (e.g., I-III), receptor status (e.g., estrogen receptor (ER), progesterone receptor (PgR), Human Epidermal Growth Factor Receptor 2 (HER2) expression status, presence of metastases, specific diagnostic codes or classifications, and specific treatment recommendations or plans, such as the type of surgery (mastectomy, breast-conserving surgery), adjuvant therapies (radiotherapy, chemotherapy, hormonal therapy), and follow-up care. It also includes details of patient history related to cancer, such as family history, previous treatments, and relevant medical conditions. \\
          \textbf{Example:} (2020-01-01: Pyyd. yst. vartijaimusolmukkeiden tutkiminen. 1.1.2020 / Doctor Doctor / RINNAN HISTOLOGINEN TUTKIMUS, LAAJA LEIKKAUSPREPARAATTI Näytenumero 0000000 RINTASYÖVÄN HISTOLOGINEN TUTKIMUS leikkaustyyppi: Osaresektaatti Kasvaimen tyypi: Duktaalinen (NST) Histologinen gradus: 1 Tubulusmuodostus (1-3): 1 Tuma-atypia (1-3): 1 Mitoosit (1-3): 1 Imutieinvaasio: Ei Ihoinvaasio: Ei Karsinomaan läpimitta (mm): 10 Lobulaarinen neoplasia: Ei.
    
    \item \textbf{Column:} osapoistot\_pat\_kertomus \\
          \textbf{Description:} osapoistot\_pat\_kertomus column contains detailed pathology reports from breast biopsies and resections, documenting findings related to breast cancer diagnosis and staging. It includes information on tumor type, presence or absence of malignancy in specified locations (e.g., in breast or lymph nodes), histological grades (e.g., I-III), receptor status (e.g., estrogen receptor (ER), progesterone receptor (PgR), Human Epidermal Growth Factor Receptor 2 (HER2) expression status, presence of metastases, specific diagnostic codes or classifications, and specific treatment recommendations or plans, such as the type of surgery (mastectomy, breast-conserving surgery), adjuvant therapies (radiotherapy, chemotherapy, hormonal therapy), and follow-up care. It also includes details of patient history related to cancer, such as family history, previous treatments, and relevant medical conditions.\\
          \textbf{Example:} (2020-01-01 12:12:12: YHTEENVETO: vasemman rinnan onkoplastisen osaresekaatin alalateralisessa osassa on 10 mm alueella on epiteeliproliferaatiota, jossa on DCIS 1-tasoinen löydös.)
    
    \item \textbf{Column:} mastektomiat\_patlaus \\
          \textbf{Description:} mastektomiat\_patlaus column contains pathology reports from breast tissue samples, primarily detailing findings from surgical procedures such as mastectomies and resections. It provides granular information, including tumor type (e.g., ductal, lobular, DCIS), histological grade, tumor size, invasive characteristics (e.g., lymphovascular invasion), margin status, lymph node status, and staging information (pTNM). These reports are crucial for diagnosis, prognosis, and treatment planning for breast cancer patients. \\
          \textbf{Example:} (2020-01-01: P. k. PAD-tutkimusta kudosnäytteistä. 10.01.2020 /Doctor Doctor / Vastaus RINNAN HISTOLOGINEN TUTKIMUS, LAAJA LEIKKAUSPREPARAATTI Näytenumero: 0000000 RINTASYÖVÄN HISTOLOGINEN TUTKIMUS Leikkaustyyppi: Mastektomia Kasvaintyyppi: DCIS Histologinen gradus: 1 Verisuoninvaasio: Ei Invaasiivisen karsinooman läpimitta (mm): 10 DCIS luokka (WHO v. 2012): 1 Marginaalit puhtaat RINTAKASVAIMEN POISTOMARGINAALIT Superiorinen/ inferiorinen (mm/mm): \textgreater30/\textgreater30 Mediaalinen/ posteriorinen (mm/mm): \textgreater30/\textgreater30 Anteriorinen/ posteriorinen (mm/mm): \textgreater30/\textgreater30 pTNM (UICC 2017): pT1bN0 Näyteenä vasemman rinnan ablaatti, kooltaan noin 30 x 20 x 10 cm.)
    
    \item \textbf{Column:} mastektomiat\_pat\_kertomus \\
          \textbf{Description:} mastektomiat\_pat\_kertomus column contains detailed pathology reports from breast cancer surgeries, primarily mastectomies. It includes information on the surgical procedure type, specimen weight, laterality, and tumor ctype, presence or absence of malignancy in specified locations (e.g., in breast or lymph nodes), histological grades (e.g., I-III), receptor status (e.g., estrogen receptor (ER), progesterone receptor (PgR), Human Epidermal Growth Factor Receptor 2 (HER2) expression status, presence of metastases, specific diagnostic codes or classifications, and details relevant to neoadjuvant therapy response, such as RCB classification and changes in tumor grade after treatment. The reports also specify staging information based on the TNM classification system.\\
          \textbf{Example:} (2020-01-01 12:12:12: Ottopvm: 01.01.2020 kysessä potilas, joille aiemmin tehty mastektomia T1 N0 reseptoripositiivisen mamma-karsinooman vuoksi.)
    
    \item \textbf{Column:} insitu\_patlaus \\
          \textbf{Description:} insitu\_patlaus column contains detailed histopathological reports for breast cancer diagnoses. It specifies tumor characteristics such as type (e.g., ductal, lobular, adenocarcinoma), grade (1–3), size (in mm), presence of DCIS (ductal carcinoma in situ) with its grade and extent, and details of invasion (lymphatic, vascular, skin) and margin status. The reports also include staging information (pTNM classification), results from sentinel lymph node biopsies (presence or absence of metastasis), and, in some cases, immunohistochemical markers (ER, PgR, Ki-67, HER2). The entries detail findings from various surgical procedures, including mastectomies, resections, and sentinel node biopsies, representing comprehensive pathological assessments crucial for treatment planning and prognosis. \\
          \textbf{Example:} (2022-01-01: DCIS-muutokset sijaitsevat hajanaisesti.), (2020-01-01: Arvioituna kasvaimen kokonaisläpimitta on 20 mm.), (2020-01-01: Marginaali DCIS:stä on posterolatelaalisesti 5 mm.)
    
    \item \textbf{Column:} insitu\_pat\_kertomus \\
          \textbf{Description:} insitu\_pat\_kertomus column contains detailed histopathological reports from breast tissue biopsies or resections, often prompted by mammographic findings or clinical suspicion. The reports document the type, presence or absence of malignancy in specified locations (e.g., in breast or lymph nodes), histological grades (e.g., I-III), receptor status (e.g., estrogen receptor (ER), progesterone receptor (PgR), Human Epidermal Growth Factor Receptor 2 (HER2) expression status, presence of metastases, and specific diagnostic codes or classifications.\\
          \textbf{Example:} (2020-01-01: P. k. PAD-tutkimusta kudosnäytteistä. 01.01.2020 / Doctor Doctor / Vastaus RINNAN HISTOLOGINEN TUTKIMUS, LAAJA LEIKKAUSPREPARAATTI Näytenumero: 0000000 RINTASYÖVÄN HISTOLOGINEN TUTKIMUS Leikkaustyyppi: Mastektomia Kasvaintyyppi: DCIS Histologinen gradus: 1 Verisuoninvaasio: Ei Invaasiivisen karsinooman läpimitta (mm): 10 DCIS luokka (WHO v. 2012): 1 Marginaalit puhtaat RINTAKASVAIMEN POISTOMARGINAALIT Superiorinen/ inferiorinen (mm/mm): \textgreater30/\textgreater30 Mediaalinen/ posteriorinen (mm/mm): \textgreater30/\textgreater30 Anteriorinen/ posteriorinen (mm/mm): \textgreater30/\textgreater30 pTNM (UICC 2017): pTisN0 Näyteenä vasemman rinnan ablaatti, kooltaan noin 30 x 20 x 10 cm.)
    
    \item \textbf{Column:} insitu\_kert\_muut \\
          \textbf{Description:} insitu\_kert\_muut column contains free-text pathology reports and clinical notes pertaining to breast cancer diagnoses and follow-up, spanning multiple dates of patient encounters. The reports document the type, presence or absence of malignancy in specified locations (e.g., in breast or lymph nodes), histological grades (e.g., I-III), receptor status (e.g., estrogen receptor (ER), progesterone receptor (PgR), Human Epidermal Growth Factor Receptor 2 (HER2) expression status, presence of metastases, and specific diagnostic codes or classifications.\
          \textbf{Example:} (2020-01-01 12:12:12: Poistetussa resekaatissa löytynyt 10 mm kokoinen invaasiivinen duktaalinen karsinooma gradus I sekä DCIS.)
    
    \item \textbf{Column:} benigni\_patlaus \\
          \textbf{Description:} benigni\_patlaus column contains histopathological reports from breast tissue samples, often following diagnostic imaging or biopsy. The reports document the type, presence or absence of malignancy in specified locations (e.g., in breast or lymph nodes), histological grades (e.g., I-III), receptor status (e.g., estrogen receptor (ER), progesterone receptor (PgR), Human Epidermal Growth Factor Receptor 2 (HER2) expression status, presence of metastases, and specific diagnostic codes or classifications.\\
          \textbf{Example:} (2020-01-01 12:12:12: Benigni.)
    
    \item \textbf{Column:} benigni\_pat\_kertomus \\
          \textbf{Description:} benigni\_pat\_kertomus column contains pathology descriptions, often derived from surgical resection specimens or biopsies related to breast conditions.\\
          \textbf{Example:} (2020-01-01 12:12:12: Benigni löydös)
    
    \item \textbf{Column:} benigni\_kert\_muut \\
          \textbf{Description:} benigni\_kert\_muut column contains free-text reports detailing various findings. It primarily describes mammography and ultrasound observations, often comparing them to previous studies.\\
          \textbf{Example:} (2020-01-01 12:12:12: Rinnassa on benigni muutos.)
    
    \item \textbf{Column:} topografia\_morfologia \\
          \textbf{Description:} topografia\_morfologia column details the specific anatomical locations and histological types of tissue samples obtained for pathological examination. It includes information on primary sites such as the breast and lymph nodes, specifying right or left sides where applicable. The column also identifies metastatic sites and provides histological diagnoses. \\
          \textbf{Example:} 2020-01-01 0000000 Female breast, right 0000000 No evidence of malignancy; 2020-01-01 Skin of breast, NOS, right 0000000 0000000 Sentinel lymph node, NOS, right 0000000 No evidence of malignancy (0/5)
    
    \item \textbf{Column:} kuuri1 \\
          \textbf{Description:} kuuri1 column contains records of the first administered (chemo)therapeutic treatment, including specific drug names, routes of administration (e.g. intravenous, subcutaneous, oral), number of cycles, and the corresponding start and end dates for each treatment course. It primarily focuses on systemic therapies used in oncology—particularly for breast cancer management—and details regimens.\\
          \textbf{Example:} Doketakseli: 1 sykli\textless e4\textgreater (2020-1-1 -- 2020-1-22)
    
    \item \textbf{Column:} kuuri2 \\
          \textbf{Description:} kuuri2 column contains records of the second administered (chemo)therapeutic treatment, including specific drug names, routes of administration (e.g. intravenous, subcutaneous, oral), number of cycles, and the corresponding start and end dates for each treatment course. It primarily focuses on systemic therapies used in oncology—particularly for breast cancer management—and details regimens.\\
          \textbf{Example:} Trastutsumabi: 1 sykli\textless e4\textgreater (2020-1-1 -- 2020-1-22)
    
    \item \textbf{Column:} kuuri3 \\
          \textbf{Description:} kuuri3 column contains records of the third administered (chemo)therapeutic treatment, including specific drug names, routes of administration (e.g. intravenous, subcutaneous, oral), number of cycles, and the corresponding start and end dates for each treatment course. It primarily focuses on systemic therapies used in oncology—particularly for breast cancer management—and details regimens.\\
          \textbf{Example:} Doketakseli: 1 sykli\textless e4\textgreater (2020-1-1 -- 2020-1-22)
    
    \item \textbf{Column:} reseptit\_laakemaaraykset \\
          \textbf{Description:} reseptit\_laakemaaraykset column contains a longitudinal record of medications prescribed to patients, denoted by ATC codes and generic drug names, alongside associated prescription dates. Entries represent individual prescriptions with start and end dates where applicable, revealing the temporal pattern of pharmacological interventions for each patient. The data suggest a broad range of therapeutic classes, including analgesics, endocrine therapies, psychotropics, and cardiovascular medications, potentially indicating the management of both cancer-related symptoms and comorbidities.\\
          \textbf{Example:} Doketakseli: 1 sykli\textless e4\textgreater (2020-1-1 -- 2020-1-22)
    
    \item \textbf{Column:} kemokur \\
          \textbf{Description:} kemokur column contains a historical record of medications administered to breast cancer patients, listing the specific drugs, their codes, and the dates of administration. It primarily details chemotherapy agents as well as targeted therapies. Additionally, supportive medications—including antiemetics and analgesics are listed.\\
          \textbf{Example:} (0000000 drug name: 2020-01-01)

    \item \textbf{Column:} leikkaukset \\
          \textbf{Description:} leikkaukset column provides a longitudinal record of surgical procedures performed on the patient, including major operations, as well as minor interventions such as diagnostic procedures. It captures the type of surgery, the anatomical location (e.g., breast, spine, lung), and often the associated surgical codes and dates, providing a comprehensive surgical history relevant to the patient’s overall medical management.\\
          \textbf{Example:} (00000 procedure name, 00000, 2020-01-01)
    
    \item \textbf{Column:} tmp \\
          \textbf{Description:} tmp column contains records of medical procedures and treatments performed on the patient, including diagnostic imaging and breast cancer-related procedures. Each record is time-stamped and encoded with a specific medical procedure or treatment code.\\
          \textbf{Example:} (00000 procedure name, 00000, 2020-01-01 12:12:12)
    
    \item \textbf{Column:} sadehoidot \\
          \textbf{Description:} sadehoidot column contains records of radiation therapy procedures, alongside dates and times of treatment delivery. The records are characterized by standardized codes indicating the type of radiation therapy.\\
          \textbf{Example:} (00000 procedure name, 00000, 2020-01-01 12:12:12)
\end{enumerate}

\clearpage
\section*{B. Question Set}
\label{supp:question_set}

This section describes the clinical questions evaluated in the Clinical Contextual Question Answering (CCQA) task. Each question corresponds to a specific clinical variable that must be inferred from the patient’s longitudinal electronic health record. For every question, predefined answer options and detailed contextual definitions were provided to ensure consistent interpretation across the annotator and models.

Clinical question types are indexed as CQ1–CQ21. Each question includes two semantically equivalent wording variants designed to assess robustness to linguistic variation: V1 (primary formulation) and V2 (rephrased formulation). Unless otherwise specified, both variants require the same answer derived from the same underlying clinical evidence.

\begin{enumerate}
    \item \textbf{CQ1 — Breast cancer first detected}: 
    \begin{enumerate}
        \item Based on the patient's record, in what year was breast cancer first detected?
        \item What year was the breast cancer initially detected?
    \end{enumerate}
    \textbf{Answer options}: 
    \begin{enumerate}
        \item None
        \item 4-Digit year (e.g., 2020)
    \end{enumerate}
    \textbf{Context}: The ground truth answer must be a single, four-digit year (YYYY) corresponding to the earliest documented diagnosis of \textbf{breast cancer}. Answer “None” if there is no documented history of breast cancer. Benign conditions and negative diagnostic results do not count as cancer.
    \vspace{3mm}

    \item \textbf{CQ2 — First diagnosis on or after 2015}:
    \begin{enumerate}
        \item Was the first breast cancer detected on or after 2015?
        \item Based on the patient’s records, was the patient’s first-ever diagnosis of breast cancer made in 2015 or later?
    \end{enumerate}
    \textbf{Answer options}: 
    \begin{enumerate}[label=\arabic*.]
        \item Yes
        \item No
    \end{enumerate}
    \textbf{Context}: Answer “Yes” if the patient’s first-ever confirmed breast cancer diagnosis occurred in 2015 or later, and there is no evidence of any earlier breast cancer diagnosis. Answer “No” if the patient had a breast cancer diagnosis prior to 2015 or has never been diagnosed with breast cancer; benign findings and negative results should also be classified as “No.”
    \vspace{3mm}

    \item \textbf{CQ3 — Breast cancer laterality}:
    \begin{enumerate}
        \item Based on the patient's record, on which side is the breast cancer located?
        \item Which breast is affected by the breast cancer?
    \end{enumerate}
    \textbf{Answer options}:
    \begin{enumerate}[label=\arabic*.]
        \item Left
        \item Right
        \item Bilateral
    \end{enumerate}
    \textbf{Context}: Breast cancer may be located in the left breast, right breast, or both breasts (referred to as bilateral breast cancer). Laterality is typically determined through clinical documentation, imaging studies (such as mammography, magnetic resonance imaging, or ultrasound), or pathology reports.
    \begin{itemize}
        \item If cancer is reported in the left breast only, select left.
        \item If cancer is reported in the right breast only, select right.
        \item If both breasts are affected, select bilateral.
        \item If the side is not documented or cannot be determined from the patient’s record, select Unknown.
    \end{itemize}
    \vspace{3mm}

    \item \textbf{CQ4 — Lesion invasiveness status}:
    \begin{enumerate}
        \item Based on the patient's record, what is the invasiveness status of the breast lesion?
        \item Is the breast cancer invasive?
    \end{enumerate}
    \textbf{Answer options}: 
    \begin{enumerate}[label=\arabic*.]
        \item DCIS
        \item Invasive
    \end{enumerate}
    \textbf{Context}:
    \begin{itemize}
        \item DCIS (ductal carcinoma in situ) refers to a non-invasive breast lesions of ductal origin. In DCIS, abnormal cells are confined within the milk ducts and have not invaded surrounding breast tissue. It is considered the earliest form of breast cancer and has not spread beyond the ductal system.
        \item Invasive breast cancer indicates that cancer cells have breached the duct or lobule walls and infiltrate surrounding breast tissue. This category includes types such invasive ductal carcinoma (IDC) i.e., carcinoma of no special type (NST), invasive lobular carcinoma (ILC), medullary carcinoma, tubular carcinoma, papillary carcinoma, mucinous carcinoma, and cribriform carcinoma. Invasive cancers have the potential to spread to lymph nodes or distant organs.
    \end{itemize}
    \vspace{3mm}

    \item \textbf{CQ5 — Histological subtype of invasive breast cancer}:
    \begin{enumerate}
        \item Based on the patient's record, what is the histological subtype of the invasive breast cancer?
        \item If the breast cancer is invasive, what is the subtype?
    \end{enumerate}
    \textbf{Answer options}:
    \begin{enumerate}[label=\arabic*.]
        \item Ductal
        \item Lobular
        \item Other
        \item Unknown
    \end{enumerate}
    \textbf{Context}: Histological subtypes of invasive breast cancer are determined by microscopic examination of tumor tissue and describe how the cancer cells are organized and appear~\cite{lukasiewicz2021breast}. The classification is based on the World Health Organization (WHO) criteria.
    \begin{itemize}
        \item \textbf{Ductal}: Arises from the epithelial cells of the breast ducts and is the most common form of invasive breast cancer. Also referred to as invasive carcinoma of no special type (NST), this diagnosis is used when the tumor lacks features of a specific histological subtype.
        \item \textbf{Lobular}: Arises from the epithelial cells of the breast lobules and is the second most common form of invasive breast cancer (invasive lobular carcinoma). It is characterized by discohesive tumor cells with diffuse infiltration, often without a discrete mass, and is more difficult to detect on mammography.
        \item \textbf{Other}: Includes less common specific histological subtypes such as medullary carcinoma, tubular carcinoma, papillary carcinoma, mucinous carcinoma, and cribriform carcinoma.
        \item \textbf{Unknown}: Select this option if the histological subtype is not documented or cannot be determined from the patient record.
    \end{itemize}
    \vspace{3mm}

    \item \textbf{CQ6 — Receipt of neoadjuvant therapy}:
    \begin{enumerate}
        \item Based on the patient's record, has the patient received neoadjuvant therapy?
        \item Was the patient a recipient of neoadjuvant therapy?
    \end{enumerate}
    \textbf{Answer options}:
    \begin{enumerate}[label=\arabic*.]
        \item Yes
        \item No
    \end{enumerate}
    \textbf{Context}:
    Neoadjuvant therapy (NAT) is any systemic cancer treatment given before surgery~\cite{baskin2024neoadjuvant,ypt}. This includes chemotherapy, immunotherapy, HER2-targeted therapy, or endocrine therapy.
    
    If the patient record shows that systemic treatment was administered before surgery, select “Yes”. If systemic treatment was given only after surgery, select “No”.
    
    Evidence of NAT in the medical record may include explicit terms such as “neoadjuvant therapy”, “neoadjuvant chemotherapy”, “preoperative treatment”, or the word “neoadjuvantti”. It may also include documentation of systemic therapy given before surgery or pathology staging using ypT or ypN, which indicates post-neoadjuvant assessment.
   \vspace{3mm}

   \item \textbf{CQ7 — Histological tumor grade}:
    \begin{enumerate}
        \item Based on the patient's record, what is the histological grade of the breast tumor?
        \item Which histological grade of breast tumor appears on the patient's record?
    \end{enumerate}
    \textbf{Answer options}:
    \begin{enumerate}[label=\arabic*.]
        \item Grade 1
        \item Grade 2
        \item Grade 3
        \item Unknown
    \end{enumerate}
    \textbf{Context}:  The histological grade of a breast tumor reflects how abnormal the cancer cells appear under a microscope and how likely the tumor is to grow or spread fast. It is determined by a pathologist using tumor tissue samples~\cite{stag_grad}.
    \begin{itemize}
        \item Grade 1/I (low grade / well-differentiated): tumor cells closely resemble normal breast cells. These tumors tend to grow slowly and are less likely to spread.
        \item Grade 2/II (intermediate grade / moderately differentiated): tumor cells show more abnormalities than Grade 1 and have a moderate growth rate.
        \item Grade 3/III (high grade / poorly differentiated): tumor cells look very different from normal cells. These cancers tend to grow and spread more aggressively.
        \item If the grade is not documented or cannot be determined from the patient’s record, select Unknown.
    \end{itemize}
    \vspace{3mm}

    \item \textbf{CQ8 — Primary tumor T classification (TNM)}:
    \begin{enumerate}
        \item Based on the patient's record, what is the T classification of the breast cancer?
        \item What breast tumor T classification is in the patient's record?
    \end{enumerate}
    \textbf{Answer options}:
    \begin{enumerate}[label=\arabic*.]
        \item Tx
        \item Tis
        \item T1
        \item T1a
        \item T1b
        \item T1c
        \item T2
        \item T3
        \item T4
        \item ypTx
        \item ypTis
        \item ypT1mi
        \item ypT1a
        \item ypT1b
        \item ypT1c
        \item ypT2
        \item ypT3
        \item ypT4
        \item Unknown
    \end{enumerate}
    \textbf{Context}: The stage of a cancer describes its size and whether it has spread from where it started. There are different ways to describe the stage of a cancer. The most commonly used one is the TNM staging system. The T classification describes the primary cancer’s size and extent. It may be determined based on clinical imaging, pathology, or post-neoadjuvant pathological assessment~\cite{TNMstaging, ypt}.
    \begin{itemize}
        \item Tis: Carcinoma in situ, including ductal carcinoma in situ (DCIS) and Paget’s disease of the nipple without invasion.
        \item T1mi: Microinvasion, defined as invasive cancer measuring 1 mm or less.
        \item T1a: Tumor larger than 0.1 cm but less than 0.5 cm.
        \item T1b: Tumor larger than 0.5 cm but less than 1.0 cm.
        \item T1c: Tumor larger than 1.0 cm but less than 2.0 cm.
        \item T2: Tumor larger than 2 cm but not larger than 5 cm.
        \item T3: Tumor larger than 5 cm.
        \item T4: Tumor has spread into the chest wall, skin, or is clinically assessed as inflammatory.
        \item Tx: Tumor size cannot be assessed.
        \item ypT: Indicates pathological post-neoadjuvant therapy tumor classification (e.g., ypT0, ypT1a).
        \item ypTx: Tumor size cannot be assessed pathologically after neoadjuvant therapy.
        \item Unknown: Use if the T classification is not available or cannot be determined from the record.
    \end{itemize}
    \vspace{3mm}
    
    \item \textbf{CQ9 — Largest invasive tumor diameter (mm)}:
    \begin{enumerate}
        \item Based on the patient's record, what is the maximum diameter of the largest breast tumor? Answer in millimeters.
        \item How wide is the breast tumor? Answer in millimeters.
    \end{enumerate}
    \textbf{Answer format}:
    \begin{enumerate}[label=\arabic*.]
        \item A numeric value between 1 and 100 (in millimeters; decimals not allowed)
        \item Unknown
    \end{enumerate}
    \textbf{Context}: The diameter of the largest invasive breast tumor refers to the greatest measurable dimension of the tumor in millimeters (mm), as assessed by pathology. This measurement corresponds to the T category in TNM staging~\cite{TNMstaging}. yp refers to post-NAT tumor size.
    \begin{itemize}
        \item Tumors $\leq$ 2 mm may be classified as (yp)T1mi (microinvasion).
        \item Tumors $>$ 2 mm and $\leq$5 mm are classified as (yp)T1a.
        \item Tumors $>$ 5 mm and $\leq$10 mm are classified as (yp)T1b.
        \item Tumors $>$ 10 mm and $\leq$20 mm are classified as (yp)T1c.
        \item Tumors $>$ 20 mm and $\leq$50 mm are classified as (yp)T2.
        \item Tumors $>50$ mm are classified as (yp)T3.
        \item If no tumor size is documented, the diameter should be marked as unknown.
    \end{itemize}
    \vspace{3mm}

    \item \textbf{CQ10 — Regional lymph node N classification (TNM)}:
    \begin{enumerate}
        \item Based on the patient's record, what is the N classification?
        \item What breast tumor N classification is in the patient's record?
    \end{enumerate}
    \textbf{Answer options}:
    \begin{enumerate}[label=\arabic*.]
        \item N0
        \item N1
        \item N0itc
        \item N1mi
        \item N1a
        \item N2a
        \item N3a
        \item NX
        \item ypN0
        \item ypN1a
        \item ypN2a
        \item ypN3a
        \item Unknown
    \end{enumerate}
    \textbf{Context}: The stage of a cancer describes its size and whether it has spread from where it started. There are different ways to describe the stage of a cancer. The most commonly used one is the TNM staging system. The N classification describes the extent of regional lymph node involvement in breast cancer. It is based on either clinical (cN) or pathologic (pN/ypN) assessment~\cite{TNMstaging, ypt}.
    \begin{itemize}
        \item N0 indicates that no regional lymph node metastasis is present.
        \item N1–N3 indicate increasing levels of nodal involvement based on the number, size, and location of metastatic nodes.
        \item ITCs (isolated tumor cells, $<$0.2 mm) fall under N0itc.
        \item N1mi refers to micrometastasis (tumor deposits $>$0.2 mm but $\leq$2 mm).
        \item NX indicates that regional lymph nodes cannot be assessed.
        \item ypN denotes nodal staging after neoadjuvant therapy (e.g., ypN0, ypN1a, ypN2a, ypN3a).
        \item Use ``Unknown" if the N classification is not documented in the patient’s record.
    \end{itemize}
    \vspace{3mm}

    \item \textbf{CQ11 — Lymph node involvement ratio}:
    \begin{enumerate}
        \item Based on the patient's record, what is the lymph node involvement ratio? Use the format [metastatic lymph nodes]/[removed lymph nodes].
        \item What is the ratio of involved lymph nodes to total lymph nodes?
    \end{enumerate}
    \textbf{Answer options}:
    \begin{enumerate}[label=\arabic*.]
        \item Unknown
        \item Numeric ratio in the format [metastatic nodes]/[removed nodes], e.g., 0/5.
    \end{enumerate}
    \textbf{Context}: The lymph node involvement ratio refers to the number of positive (cancer-involved) lymph nodes out of the total number of lymph nodes examined, typically after surgical removal~\cite{zhou2021progression}. It is expressed in the format: [positive nodes]/[total nodes]. This ratio helps assess the extent of regional spread and is routinely reported in breast cancer pathology. If lymph node data are unavailable or not documented, the value should be recorded as unknown~\cite{zhou2021progression}.
    \vspace{3mm}

    \item \textbf{CQ12 — Presence of distant metastasis}:
    \begin{enumerate}
        \item Based on the patient's record, does the patient have metastatic disease?
        \item Is there evidence of metastatic breast cancer in the patient's record?
    \end{enumerate}
    \textbf{Answer options}:
    \begin{enumerate}[label=\arabic*.]
        \item Yes
        \item No
        \item Unknown
    \end{enumerate}
    \textbf{Context}: Metastatic breast cancer is defined by the presence of distant spread beyond the breast and regional lymph nodes. Common metastatic sites include the bone, liver, lungs, and brain. Detection typically involves imaging (CT, MRI, PET-CT, or bone scan) and may be supported by clinical symptoms or biopsy confirmation. Imaging is generally recommended in patients with advanced-stage tumors (T3/T4), multiple involved lymph nodes, aggressive subtypes, or symptoms suggestive of metastasis (e.g., bone pain, abnormal liver function tests). If the record does not indicate confirmed distant spread, metastatic status should be marked as unknown~\cite{pesapane2020imaging}.
    \vspace{3mm}
    
    \item \textbf{CQ13 — Estrogen receptor (ER) expression percentage}:
    \begin{enumerate}
        \item Based on the patient's record, what is the estrogen receptor (ER) expression percentage?
        \item What is the patient's percentage of estrogen receptor expression?
    \end{enumerate} 
    \textbf{Answer format}:  
    \begin{enumerate}[label=\arabic*.]
        \item An integer between 0 and 100 (representing percentage)
        \item Unknown
    \end{enumerate}
    \textbf{Context}: Estrogen receptor (ER) expression is determined using immunohistochemistry (IHC) and reported as the percentage of tumor cell nuclei that stain positive for ER. This percentage reflects the proportion of cancer cells expressing estrogen receptors and is recorded as an integer between 0 and 100. ER expression is routinely evaluated in breast cancer and plays a critical role in treatment planning. If the ER percentage is not reported in the patient's record, the value should be marked as unknown~\cite{hrs, hrsreport}.
    \vspace{3mm}

    \item \textbf{CQ14 — Progesterone receptor (PR/PgR) expression percentage}:
    \begin{enumerate}
        \item Based on the patient's record, what is the progesterone receptor (PR/PgR) expression percentage?
        \item What is the patient's percentage of progesterone receptor expression?
    \end{enumerate}
    \textbf{Answer format}:  
    \begin{enumerate}[label=\arabic*.]
        \item An integer between 0 and 100 (representing percentage)
        \item Unknown
    \end{enumerate}
    \textbf{Context}: Progesterone receptor (PR) expression is assessed by immunohistochemistry (IHC) and reported as the percentage of tumor cell nuclei that stain positive for PR/PgR. The result reflects how many cancer cells express PR/PgR and is recorded as an integer between 0 and 100. PR/PgR expression is routinely reported in breast cancer pathology and helps guide treatment decisions. If no percentage is reported in the patient’s record, the value should be marked as unknown~\cite{hrsreport}.
    \vspace{3mm}

    \item \textbf{CQ15 — Ki-67 proliferation index percentage}:
    \begin{enumerate}
        \item Based on the patient's record, what is the Ki-67 proliferation index percentage?\
        \item What is the patient's percentage of Ki-67 proliferation index?
    \end{enumerate}
    \textbf{Answer format}:  
    \begin{enumerate}[label=\arabic*.]
        \item An integer between 0 and 100 (representing percentage)
        \item Unknown
    \end{enumerate}
    \textbf{Context}: Ki-67 is a protein found in the nucleus of dividing cancer cells. The Ki-67 proliferation index is reported as a percentage, representing the proportion of tumor cells actively dividing. In breast cancer, a higher Ki-67 percentage suggests a more aggressive tumor. Results are typically expressed as an integer from 0 to 100. If no Ki-67 value is available, the status is recorded as unknown~\cite{ki67}.
    \vspace{3mm}

    \item \textbf{CQ16 — HER2 expression status}:
    \begin{enumerate}
        \item Based on the patient’s record, what is the HER2 expression status?
        \item Has the patient HER2 positive or negative tumor?
    \end{enumerate}
    \textbf{Answer options}:
    \begin{enumerate}[label=\arabic*.]
        \item Positive
        \item Negative
        \item Unknown
    \end{enumerate}
    \textbf{Context}: HER2 (human epidermal growth factor receptor 2) is a protein involved in cell growth. Immunohistochemistry (IHC) is a test which detects the amount of HER2 protein on the surface of the cancer cells. Breast cancers are classified as HER2-positive if they show high levels of HER2 protein (IHC 3+) or have HER2 gene amplification (positive FISH). Cancers with no or low HER2 expression (IHC 0, 1+, or 2+ with negative FISH) are considered HER2-negative~\cite{hrsreport}.
    \vspace{3mm}

    \item \textbf{CQ17 — Triple-negative breast cancer (TNBC) status}:
    \begin{enumerate}
        \item Based on the patient’s record, is the breast cancer triple-negative?
        \item Does the patient have triple-negative breast cancer?
    \end{enumerate}
    \textbf{Answer options}:
    \begin{enumerate}[label=\arabic*.]
        \item Yes
        \item No
        \item Unknown
    \end{enumerate}
    \textbf{Context}: Triple-negative breast cancer (TNBC) is defined as a tumor that tests negative for estrogen receptors (ER), progesterone receptors (PR), and HER2. If all three markers are negative, the cancer is considered triple-negative. If any of the three markers are positive, the tumor is not classified as triple-negative~\cite{triplenegative}.
    \vspace{3mm}
    
    \item \textbf{CQ18 — Receipt of adjuvant chemotherapy}:
    \begin{enumerate}
        \item Based on the patient’s record, has the patient received adjuvant chemotherapy? 
        \item Was the patient a recipient of adjuvant chemotherapy?
    \end{enumerate}
    \textbf{Answer options}:
    \begin{enumerate}[label=\arabic*.]
        \item Yes
        \item No
        \item Unknown
    \end{enumerate}
    \textbf{Context}: Adjuvant chemotherapy is given after surgery to reduce the risk of cancer recurrence by targeting microscopic disease that may remain in the body. It is recommended based on tumor characteristics—such as size, grade, lymph node involvement, and receptor status—as well as patient factors like age, overall health, and personal preference. Unlike in metastatic disease, there is no visible tumor response to assess~\cite{jones2016cytotoxic}.
    \vspace{3mm}

    \item \textbf{CQ19 — Receipt of adjuvant radiation therapy}:
    \begin{enumerate}
        \item Based on the patient’s record, has the patient received adjuvant radiation therapy?
        \item Was the patient a recipient of adjuvant radiation therapy?
    \end{enumerate}
    \textbf{Answer options}:
    \begin{enumerate}[label=\arabic*.]
        \item Yes
        \item No
        \item Unknown
    \end{enumerate}
    \textbf{Context}: Adjuvant therapy is treatment given after surgery to reduce the risk of cancer coming back. One common form is adjuvant radiation therapy, which uses high-energy beams (usually external beam radiation) to destroy any remaining cancer cells in or near the breast or chest wall after surgery. This is often recommended for patients who had breast-conserving surgery or have higher risk of recurrence~\cite{radiationtherapy}. 
    
    In the clinical records, information about this treatment might be explicitly mentioned: ``Adjuvanttisädehoito".
    \vspace{3mm}
    
    \item \textbf{CQ20 — Receipt of adjuvant hormonal therapy}:
    \begin{enumerate}
        \item Based on the patient’s record, has the patient received adjuvant hormonal therapy?
        \item Was the patient a recipient of adjuvant hormonal therapy?
    \end{enumerate}
    \textbf{Answer options}:
    \begin{enumerate}[label=\arabic*.]
        \item Yes
        \item No
        \item Unknown
    \end{enumerate}
    \textbf{Context}: Adjuvant therapy is treatment given after primary treatment, such as surgery, to reduce the risk of cancer recurrence. In hormone receptor–positive breast cancer, adjuvant hormonal therapy helps block or lower estrogen and/or progesterone to prevent cancer growth~\cite{radiationtherapy}.
    
    Common hormonal therapies include tamoxifen and aromatase inhibitors like anastrozole, letrozole, and exemestane. These are typically used in patients who have estrogen (ER) or progesterone (PR/PgR) receptor positive breast cancer ~\cite{radiationtherapy}.

    In the clinical records, information about this treatment might be explicitly mentioned: ``Adjuvanttihormonihoito".
    \vspace{3mm}

    \item \textbf{CQ21 — Receipt of adjuvant anti-HER2 therapy}:
    \begin{enumerate}
        \item Based on the patient’s record, has the patient received adjuvant anti-HER2 therapy?
        \item Was the patient a recipient of adjuvant anti-HER2 therapy?
    \end{enumerate}
    \textbf{Answer options}:
    \begin{enumerate}[label=\arabic*.]
        \item Yes
        \item No
        \item Unknown
    \end{enumerate}
    \textbf{Context}: Multiple effective therapies now target the HER2 pathway using different mechanisms, acting either inside or outside the cell. Since the approval of trastuzumab in 2001—the first HER2-targeted therapy—several additional agents have been developed and evaluated in both metastatic and (neo)adjuvant settings in breast cancer. These include lapatinib, pertuzumab, neratinib, and T-DM1 ~\cite{debiasi2018efficacy}.
    
    In the adjuvant setting, trastuzumab has long been the cornerstone of treatment for HER2-positive breast cancer. More recently, combinations have been approved based on improved outcomes in selected patients~\cite{debiasi2018efficacy}.
    
    In the clinical records, information about this treatment might be explicitly mentioned: "Anti-HER2".
    
\end{enumerate}

\clearpage
\section*{C. Prompt Template and Construction}
\label{supp:prompt_template}

All large language models were evaluated using a unified role-based prompt format. Prompts were specified as structured messages with explicit \texttt{system} and \texttt{user} roles and converted to model-specific input text using the Hugging Face \texttt{apply\_chat\_template} function prior to generation with \texttt{model.generate()}. This approach ensured consistent instructions and input structure across models.

Each prompt consisted of two messages: (i) a \texttt{system} message containing general task instructions, optionally augmented with task-specific contextual information, and (ii) a \texttt{user} message containing the patient report, the question, and the allowed answer options.

\subsection*{System message (instructions)}
The base instruction text was:

\begin{quote}\small
\texttt{You are an expert mammography radiologist analyzing a mammography report. Answer the provided question based solely on the report's details. Select the single most accurate option from the given choices. One option is always correct. Respond only with the selected answer (e.g., <first option>) without explanations.}
\end{quote}

If \texttt{task\_context=True}, the system message was extended with the curated \textit{Context} text associated with the current question (see Section~B).

\subsection*{User message (content)}
The user message contained four fixed blocks:
\begin{enumerate}
    \item \textbf{Patient Medical Report:} the patient-specific EHR record, constructed by concatenating all CSV fields into a single text string, with line breaks separating individual fields.
    \item \textbf{Question:} the task-specific question.
    \item \textbf{Answer Options:} a bullet list of all permitted answer strings, listed in alphabetical order to ensure a consistent, model-independent ordering.
    \item \textbf{Correct Answer:} a placeholder line containing ``\verb|[Your answer here]|'', which remained empty during inference.
\end{enumerate}

\subsection*{One-shot variant}
If \texttt{one\_shot=True}, the prompt began with an \textit{Example} block consisting of a sample patient report, its corresponding question, the full set of answer options, and the correct answer. This example preceded the \textit{Task} block, which contained the target patient report and an empty answer field. The example was drawn from data not included in the evaluation set.

\subsection*{Exact string layout}

For the standard (non–one-shot) configuration, the user message followed a fixed structured format comprising the patient record, the clinical question, and predefined answer options. The template was defined as follows:

\begin{quote}\small
\begin{verbatim}
**Patient Medical Report:**
<key_1>: <value_1>
...
<key_k>: <value_k>

**Question:**
<question text>

**Answer Options:**
- <option A>
- <option B>
...

**Correct Answer:**
[Your answer here]
\end{verbatim}
\end{quote}

\subsection*{Chat serialization}

Prior to model inference, prompts were serialized into a two-turn message structure consisting of a system message containing task instructions and a user message containing the input data. The serialized format was defined as follows:

\begin{verbatim}
[
  {"role": "system",
   "content": <INSTRUCTIONS AS ABOVE>
  },
  {"role": "user",
   "content": <CONTENT AS ABOVE>
  }
]
\end{verbatim}

\clearpage
\section*{D. Prompt Conditioning Experiments}
\label{supp:prompt_conditioning}

One-shot prompting and extended context conditioning were evaluated to assess whether additional task-specific guidance improves clinical contextual question answering performance. Accuracy results for both strategies are shown in Supplementary Table~\ref{tab:conditioning}. Conditioning produced only modest and inconsistent effects across models, with small improvements observed for some architectures and negligible changes for others. Models marked with $^{\dagger}$ were not evaluated under these strategies because the additional prompt content exceeded their supported context window. These findings are consistent with the main Results, indicating that performance on this task is driven primarily by intrinsic model capability rather than prompt design.

\begin{table*}[ht]
\centering
\footnotesize
\caption{\textbf{Conditioning strategies for clinical contextual question answering.}
Accuracy is reported for models evaluated in BF16 precision under one-shot prompting and extended context conditioning. $\Delta$ denotes absolute change in accuracy relative to baseline, expressed in percentage points. Models marked with $^{\dagger}$ were not evaluated under these strategies because the additional prompt content exceeded their supported context window. All other models were evaluated on the full dataset ($N=1{,}664$).}
\label{tab:conditioning}
\begin{tabular}{llcccc}
\toprule
\multirow{2}{*}{\textbf{Developer}} &
\multirow{2}{*}{\textbf{Model}} &
\multicolumn{2}{c}{\textbf{1-shot}} &
\multicolumn{2}{c}{\textbf{Context}} \\
\cmidrule(lr){3-4}\cmidrule(lr){5-6}
 &  & \textbf{Acc.} & $\boldsymbol{\Delta}$ & \textbf{Acc.} & $\boldsymbol{\Delta}$ \\
\midrule
\multirow{4}{*}{\textbf{Google}}
 & Gemma-3-4B-it~\cite{2025gemma3}  & 69.5 & $-3.1$ & 72.5 & $-0.1$ \\
 & Gemma-3-27B-it~\cite{2025gemma3} & 94.0 & $+1.3$ & 94.1 & $+1.4$ \\
 & MedGemma-4B-it~\cite{2025medgemma} & 70.4 & $-4.1$ & 73.9 & $-0.6$ \\
 & MedGemma-27B-text-it~\cite{2025medgemma} & 94.0 & $+1.9$ & 93.7 & $+1.6$ \\
\addlinespace
\multirow{3}{*}{\textbf{DeepSeek}}
 & R1-Distill-Qwen-7B~\cite{2025deepseek-r1} & 50.3 & $-0.1$ & N/A & N/A \\
 & R1-Distill-Qwen-32B~\cite{2025deepseek-r1} & 92.3 & $-0.4$ & N/A & N/A \\
 & R1-Distill-Llama-70B~\cite{2025deepseek-r1} & 93.3 & $-1.5$ & N/A & N/A \\
\addlinespace
\multirow{3}{*}{\textbf{Meta}}
 & Llama-3.1-8B~\cite{2024llama3}  & 79.9 & $0.0$ & 80.3 & $+0.4$ \\
 & Llama-3.1-70B~\cite{2024llama3}  & \textbf{95.4} & $+0.1$ & \textbf{95.7} & $+0.4$ \\
 & Llama-3.3-70B~\cite{2024llama3}  & 94.5 & $-0.4$ & 95.0 & $+0.1$ \\
\addlinespace
\multirow{2}{*}{\textbf{Qwen}}
 & Qwen3-4B-2507~\cite{yang2025qwen3} & 92.3 & $+1.2$ & 91.2 & $+0.1$ \\
 & Qwen3-30B-A3B-2507~\cite{yang2025qwen3} & 93.8 & $-1.3$ & 95.1 & $0.0$ \\
\addlinespace
\multirow{2}{*}{\textbf{LumiOpen}}
 & Llama-Poro-2-8B~\cite{poro2_2025}$^{\dagger}$ & -- & -- & -- & -- \\
 & Llama-Poro-2-70B~\cite{poro2_2025}$^{\dagger}$ & -- & -- & -- & -- \\
\bottomrule
\end{tabular}
\end{table*}

\clearpage
\section*{E. Per-question accuracy}
\label{supp:per_question_accuracy}

Per-question accuracy varied substantially across clinical questions CQ1--CQ21 for the primary wording variant (V1), as shown in Supplementary Tables~\ref{tab:supp_perq_q1_q11} and~\ref{tab:supp_perq_q12_q21}.

Mean accuracy across models ranged from 62.8\% for CQ6 to 93.7\% for CQ3. Several questions demonstrated consistently high performance across models, including CQ3--CQ5, CQ7, CQ16, CQ18, and CQ21, each with mean accuracy between 88.9\% and 93.7\%. In contrast, CQ6 showed the lowest performance and the widest variation across models, with accuracy ranging from 4.7\% to 98.7\%. CQ14 also exhibited comparatively low performance and substantial variability, ranging from 25.0\% to 95.7\%. Intermediate performance was observed for CQ8, CQ9, CQ11, and CQ13, with moderate inter-model variability.

These findings indicate marked differences in task difficulty across clinical questions, suggesting that aggregate benchmark scores may obscure clinically relevant weaknesses in specific information extraction tasks.

\begin{sidewaystable}
\centering
\normalsize
\caption{\textbf{Per-question accuracy (\%) by model for clinical question types 1--11.}
Entries report accuracy for the primary wording variant (V1) of each clinical question.
Models marked with $^{\dagger}$ were evaluated on a context-limited subset ($\leq$8,192 tokens, $N=955$) and are not directly comparable to long-context models. Bold indicates the highest accuracy for each clinical question among fully evaluated models (i.e., excluding $^{\dagger}$ models; ties jointly bolded). Keyword matching (regex) is a deterministic rule-based baseline using regular-expression matching.}
\label{tab:supp_perq_q1_q11}
\begin{tabular}{llccccccccccc}
\toprule
\multirow{2}{*}{\textbf{Developer}} & \multirow{2}{*}{\textbf{Model}} &
\multicolumn{11}{c}{\textbf{Clinical question ID}} \\
\cmidrule(lr){3-13}
& & CQ1 & CQ2 & CQ3 & CQ4 & CQ5 & CQ6 & CQ7 & CQ8 & CQ9 & CQ10 & CQ11 \\
 &  & N=168 & N=176 & N=82 & N=85 & N=71 & N=75 & N=79 & N=73 & N=62 & N=78 & N=71 \\
\midrule
\multirow{4}{*}{Google}
& Gemma-3-4B-it         & 88.1 & 56.2 & 86.6 & 87.1 & 94.4 & 9.3 & 88.6 & 37.0 & 77.4 & 75.6 & 73.2 \\
& Gemma-3-27B-it        & 94.6 & \textbf{98.9} & \textbf{100} & \textbf{100} & \textbf{95.8} & 74.7 & 98.7 & 91.8 & 79.0 & 96.2 & 95.8 \\
& MedGemma-4B-it        & 85.1 & 55.1 & 85.4 & 96.5 & 93.0 & 41.3 & 87.3 & 11.0 & 77.4 & 93.6 & 60.6 \\
& MedGemma-27B-text-it  & 96.4 & 91.5 & \textbf{100} & \textbf{100} & \textbf{95.8} & 82.7 & \textbf{100} & 95.9 & 71.0 & \textbf{97.4} & \textbf{98.6} \\
\addlinespace
\multirow{3}{*}{DeepSeek}
& R1-Distill-Qwen-7B    & 44.0 & 70.5 & 58.5 & 40.0 & 69.0 & 41.3 & 63.3 & 43.8 & 35.5 & 51.3 & 12.7 \\
& R1-Distill-Qwen-32B   & 91.7 & 96.6 & 97.6 & 96.5 & \textbf{95.8} & 78.7 & 98.7 & \textbf{98.6} & 75.8 & 94.9 & 90.1 \\
& R1-Distill-Llama-70B  & \textbf{97.6} & 97.7 & 98.8 & 97.6 & 94.4 & 85.3 & 98.7 & 95.9 & 72.6 & 94.9 & 95.8 \\
\addlinespace
\multirow{3}{*}{Meta}
& Llama-3.1-8B          & 94.6 & 70.5 & 97.6 & 97.6 & 93.0 & 10.7 & 87.3 & 87.7 & 75.8 & 94.9 & 74.6 \\
& Llama-3.1-70B         & 96.4 & 98.3 & \textbf{100} & \textbf{100} & 90.1 & 92.0 & 98.7 & 97.3 & 83.9 & 94.9 & 95.8 \\
& Llama-3.3-70B         & 96.4 & \textbf{98.9} & \textbf{100} & \textbf{100} & 93.0 & 89.3 & 98.7 & 97.3 & 79.0 & 93.6 & 95.8 \\
\addlinespace
\multirow{2}{*}{Qwen}
& Qwen3-4B-2507         & 90.5 & 94.9 & \textbf{100} & 96.5 & 93.0 & 86.7 & 96.2 & 94.5 & 83.9 & 96.2 & 70.4 \\
& Qwen3-30B-A3B-2507    & 96.4 & 97.2 & \textbf{100} & \textbf{100} & 93.0 & \textbf{98.7} & \textbf{100} & 97.3 & \textbf{87.1} & 92.3 & 93.0 \\
\addlinespace
\multirow{2}{*}{LumiOpen}
& Llama-Poro-2-8B$^{\dagger}$  & 79.5 & 80.0 & 87.0 & 100 & 71.8 & 4.7 & 75.0 & 76.9 & 72.2 & 69.0 & 58.8 \\
& Llama-Poro-2-70B$^{\dagger}$ & 89.3 & 91.2 & 100 & 93.8 & 89.7 & 55.8 & 93.2 & 84.6 & 94.4 & 92.9 & 97.1 \\
\addlinespace
Baseline & Rule-based (substring) & 48.2 & 53.4 & 67.1 & 15.3 & 85.9 & 4.0 & 59.5 & 37.0 & 4.8 & 83.3 & 90.1  \\
\midrule
\multicolumn{2}{l}{\textbf{Mean across models (LLMs only)}} & 88.8 & 85.5 & 93.7 & 93.0 & 90.8 & 62.8 & 92.3 & 79.1 & 75.6 & 88.9 & 79.6 \\
\bottomrule
\end{tabular}
\end{sidewaystable}

\begin{sidewaystable}
\centering
\normalsize
\caption{\textbf{Per-question accuracy (\%) by model for clinical question types 12--21.}
Entries report accuracy for the primary wording variant (V1) for each question.
Models marked with $^{\dagger}$ were evaluated on a context-limited subset ($\leq$8,192 tokens, $N=955$) and are not directly comparable to long-context models. Bold indicates the highest accuracy for each clinical question among fully evaluated models (i.e., excluding $^{\dagger}$ models; ties jointly bolded). Keyword matching (regex) is a deterministic rule-based baseline using regular-expression matching.}
\label{tab:supp_perq_q12_q21}
\begin{tabular}{llcccccccccc}
\toprule
\multirow{2}{*}{\textbf{Developer}} &
\multirow{2}{*}{\textbf{Model}} &
\multicolumn{10}{c}{\textbf{Clinical question ID}} \\
\cmidrule(lr){3-12}
& & CQ12 & CQ13 & CQ14 & CQ15 & CQ16 & CQ17 & CQ18 & CQ19 & CQ20 & CQ21 \\
 &  & N=78 & N=45 & N=46 & N=51 & N=61 & N=72 & N=71 & N=75 & N=72 & N=73 \\
\midrule
\multirow{4}{*}{Google}
& Gemma-3-4B-it         & 73.1 & 66.7 & 47.8 & 92.2 & 93.4 & 61.1 & 88.7 & 86.7 & 72.2 & 67.1 \\
& Gemma-3-27B-it        & 76.9 & 84.4 & 58.7 & 86.3 & 98.4 & \textbf{100} & 95.8 & \textbf{97.3} & 93.1 & \textbf{100} \\
& MedGemma-4B-it        & \textbf{96.2} & 55.6 & 47.8 & 92.2 & 95.1 & 66.7 & 87.3 & 89.3 & 73.6 & 71.2 \\
& MedGemma-27B-text-it  & 79.5 & 71.1 & 63.0 & 70.6 & 98.4 & \textbf{100} & 95.8 & 96.0 & 97.2 & \textbf{100} \\
\addlinespace
\multirow{3}{*}{DeepSeek}
& R1-Distill-Qwen-7B    & 71.8 & 28.9 & 26.1 & 41.2 & 47.5 & 44.4 & 59.2 & 53.3 & 56.9 & 53.4 \\
& R1-Distill-Qwen-32B   & 74.4 & 91.9 & 91.3 & \textbf{98.0} & 98.4 & \textbf{100} & 94.4 & 88.0 & 91.7 & 98.6 \\
& R1-Distill-Llama-70B  & 73.1 & \textbf{93.3} & \textbf{95.7} & \textbf{98.0} & \textbf{100} & 98.6 & \textbf{98.6} & 96.0 & \textbf{98.6} & \textbf{100} \\
\addlinespace
\multirow{3}{*}{Meta}
& Llama-3.1-8B          & 61.5 & 37.8 & 76.1 & 92.2 & \textbf{100} & 66.7 & 85.9 & 90.7 & 72.2 & 89.0 \\
& Llama-3.1-70B         & 78.3 & 84.4 & 91.3 & \textbf{98.0} & \textbf{100} & \textbf{100} & 97.2 & 94.7 & \textbf{98.6} & \textbf{100} \\
& Llama-3.3-70B         & 71.8 & 88.9 & \textbf{95.7} & 96.1 & \textbf{100} & \textbf{100} & 97.2 & 94.7 & 95.8 & \textbf{100} \\
\addlinespace
\multirow{2}{*}{Qwen}
& Qwen3-4B-2507         & 80.8 & 84.4 & 65.2 & 94.1 & 95.1 & 98.6 & 97.2 & 89.3 & 88.9 & 98.6 \\
& Qwen3-30B-A3B-2507    & 69.2 & \textbf{93.3} & 82.6 & \textbf{98.0} & \textbf{100} & \textbf{100} & \textbf{100} & 94.7 & 94.4 & \textbf{100} \\
\addlinespace
\multirow{2}{*}{LumiOpen}
& Llama-Poro-2-8B$^{\dagger}$  & 78.0 & 33.3 & 25.0 & 85.2 & 74.2 & 57.5 & 59.0 & 85.4 & 71.8 & 58.5 \\
& Llama-Poro-2-70B$^{\dagger}$ & 87.8 & 87.5 & 70.8 & 96.3 & 100 & 95.0 & 100 & 92.7 & 87.2 & 100 \\
\addlinespace
Baseline & Rule-based (substring) & 0.0 & 11.1 & 8.7 & 3.9 & 78.7 & 16.7 & 62.0 & 88.0 & 66.7 & 17.8 \\
\midrule
\multicolumn{2}{l}{\textbf{Mean across models (LLMs only)}} & 76.1 & 72.3 & 68.3 & 88.3 & 93.3 & 85.5 & 90.4 & 89.2 & 85.6 & 88.9 \\
\bottomrule
\end{tabular}
\end{sidewaystable}

\clearpage
\section*{F. Performance as a Function of EHR Length}
\label{supp:per_length_accuracy}

Accuracy varied with electronic health record (EHR) length, as summarized in Supplementary Table~\ref{tab:length_quartiles}. Records were stratified into quartiles spanning 1{,}420–26{,}739 tokens (Q1–Q4, shortest to longest). Sensitivity to increasing EHR length varied across models. For example, Llama-3.1-70B, Llama-3.3-70B, and Qwen3-30B-A3B-2507 showed only small decreases in accuracy across input lengths. In contrast, some models exhibited greater sensitivity to record length, with R1-Distill-Qwen-7B showing a pronounced decline across quartiles. Poro-2 models, with a nominal maximum input length of 8{,}192 tokens, showed substantial deterioration on longer records, indicating reduced performance when inputs exceeded the context lengths for which the model was optimized. Because quartiles differed in both input length and underlying patient cases, these results reflect combined effects of record length and case composition and complexity rather than a purely causal effect of length alone.

\begin{table*}[ht]
\centering
\footnotesize
\caption{\textbf{Clinical contextual question answering accuracy by electronic health record length quartile.}
Records were stratified into quartiles based on input length measured in tokens using the Meta Llama tokenizer as a fixed reference (Q1 shortest, Q4 longest; ranges: Q1 1{,}420–5{,}808, Q2 5{,}808–8{,}197, Q3 8{,}197–11{,}635, Q4 11{,}635–26{,}739 tokens). Token counts may vary slightly across models because of tokenizer differences. Values report accuracy for the primary question variant (V1). Results marked with $^{\dagger}$ were evaluated only on records that fully fit within the supported context window ($\leq$8{,}192 tokens, $N=955$) and are not directly comparable across all quartiles. Bold indicates the highest accuracy among directly comparable models within each quartile.}
\label{tab:length_quartiles}
\begin{tabular}{llcccc}
\toprule
\multirow{2}{*}{\textbf{Developer}} &
\multirow{2}{*}{\textbf{Model}} &
\multicolumn{4}{c}{\textbf{V1 Acc. (\%)}} \\
\cmidrule(lr){3-6}
 &  & \textbf{Q1} & \textbf{Q2} & \textbf{Q3} & \textbf{Q4} \\
\midrule
\multirow{4}{*}{Google}
 & Gemma-3-4B-it~\cite{2025gemma3} & 67.9 & 73.1 & 73.0 & 76.2 \\
 & Gemma-3-27B-it~\cite{2025gemma3} & 94.6 & 93.4 & 93.5 & 89.3 \\
 & MedGemma-4B-it~\cite{2025medgemma} & 70.9 & 71.1 & 77.0 & 78.2 \\
 & MedGemma-27B-text-it~\cite{2025medgemma} & 94.1 & 93.4 & 91.6 & 89.6 \\
\addlinespace
\multirow{3}{*}{DeepSeek}
 & R1-Distill-Qwen-7B~\cite{2025deepseek-r1} & 61.3 & 56.1 & 48.3 & 40.6 \\
 & R1-Distill-Qwen-32B~\cite{2025deepseek-r1} & 94.7 & 93.2 & 92.2 & 91.3 \\
 & R1-Distill-Llama-70B~\cite{2025deepseek-r1} & \textbf{96.9} & \textbf{95.5} & 93.9 & 91.8 \\
\addlinespace
\multirow{3}{*}{Meta}
 & Llama-3.1-8B~\cite{2024llama3}  & 78.5 & 80.9 & 84.6 & 74.8 \\
 & Llama-3.1-70B~\cite{2024llama3}  & 96.4 & 94.9 & \textbf{95.8} & 93.8 \\
 & Llama-3.3-70B~\cite{2024llama3}  & 96.6 & 94.9 & 94.8 & 92.6 \\
\addlinespace
\multirow{2}{*}{Qwen}
 & Qwen3-4B-2507~\cite{yang2025qwen3} & 92.0 & 91.9 & 90.6 & 90.3 \\
 & Qwen3-30B-A3B-2507~\cite{yang2025qwen3} & 96.3 & 95.2 & 94.4 & \textbf{94.7} \\
\addlinespace
\multirow{2}{*}{LumiOpen}
 & Llama-Poro-2-8B~\cite{poro2_2025}$^{\dagger}$ & 74.5 & 65.8 & 36.6 & 0.9 \\
 & Llama-Poro-2-70B~\cite{poro2_2025}$^{\dagger}$ & 92.6 & 89.1 & 72.4 & 6.2 \\
\bottomrule
\end{tabular}
\end{table*}

\clearpage
\section*{G. Statistical analysis}
\label{supp:statistical}

Statistical uncertainty of model accuracy was quantified using 95\% confidence intervals computed with the exact Clopper–Pearson method for binomial proportions (Supplementary Table~\ref{tab:supp_accuracy_ci}). Pairwise statistical comparisons between models were performed using paired McNemar tests applied to the binary correctness outcomes for each patient–question instance under the Version~1 (V1) question formulation. Exact binomial tests were used to compute two-sided $P$ values based on the number of discordant instances ($n_{10}$ and $n_{01}$). To account for multiple comparisons, $P$ values were adjusted using the Holm correction, and the resulting comparisons are reported in Supplementary Table~\ref{tab:supp_mcnemar}.

\begin{table*}[ht]
\centering
\footnotesize
\caption{\textbf{Clinical contextual question answering accuracy with 95\% confidence intervals.} Accuracy is reported for Version~1 (V1) and Version~2 (V2) questions, as well as consistency across variants. Confidence intervals (CI) were computed using the exact Clopper–Pearson method. Bold indicates the best result. OOM indicates models that exceeded the available GPU memory and could not be evaluated. Models marked with $^{\dagger}$ were evaluated on a reduced subset ($N=955$). All other models were evaluated on the full dataset ($N=1{,}664$).}
\label{tab:supp_accuracy_ci}
\begin{tabular}{llccc}
\toprule
\multirow{2}{*}{\textbf{Developer}} & \multirow{2}{*}{\textbf{Model}} & \textbf{V1 accuracy} & \textbf{V2 accuracy} & \textbf{Consistency (V1--V2)} \\
\cmidrule(lr){3-5}
  & & \textbf{\% (95\% CI)} & \textbf{\% (95\% CI)} & \textbf{\% (95\% CI)} \\
\midrule
\multirow{4}{*}{Google}
 & Gemma-3-4B-it~\cite{2025gemma3}  & 72.6 (70.4–74.7) & 68.2 (65.9–70.4) & 83.2 (81.3–85.0) \\
 & Gemma-3-27B-it~\cite{2025gemma3}  & 92.7 (91.4–93.9) & 92.8 (91.4–94.0) & 96.3 (95.3–97.2) \\
 & MedGemma-4B-it~\cite{2025medgemma} & 74.5 (72.3–76.5) & 73.1 (70.9–75.3) & 89.6 (88.0–91.0) \\
 & MedGemma-27B-text-it~\cite{2025medgemma} & 92.1 (90.7–93.4) & 90.0 (88.4–91.4) & 94.0 (92.7–95.1) \\
\addlinespace
\multirow{3}{*}{DeepSeek}
 & R1-Distill-Qwen-7B~\cite{2025deepseek-r1} & 50.4 (47.9–52.8) & 51.2 (48.8–53.6) & 43.4 (41.1–45.9) \\
 & R1-Distill-Qwen-32B~\cite{2025deepseek-r1} & 92.7 (91.3–93.9) & 94.6 (93.4–95.6) & 94.7 (93.5–95.7) \\
 & R1-Distill-Llama-70B~\cite{2025deepseek-r1} & 94.8 (93.6–95.8) & \textbf{95.1} (93.9–96.1) & 97.0 (96.1–97.8) \\
\addlinespace
\multirow{4}{*}{Meta}
 & Llama-3.1-8B~\cite{2024llama3} & 79.9 (77.9–81.8) & 80.2 (78.2–82.1) & 94.2 (92.9–95.2) \\
 & Llama-3.1-70B~\cite{2024llama3} & \textbf{95.3} (94.2–96.6) & 94.2 (93.0–95.3) & 97.3 (96.4–98.0) \\
 & Llama-3.3-70B~\cite{2024llama3} & 94.9 (93.7–95.9) & 94.9 (93.7–95.9) & 97.9 (97.1–98.5) \\
 & Llama-4-Scout-17B~\cite{MetaAI2025Llama4} & OOM & OOM & — \\
\addlinespace 
OpenAI & GPT-OSS-20B~\cite{openai2025gptoss120bgptoss20bmodel} & OOM & OOM & — \\
\addlinespace
\multirow{2}{*}{Qwen}
 & Qwen3-4B-2507~\cite{yang2025qwen3} & 91.1 (89.6–92.4) & 89.8 (88.2–91.2) & 95.3 (94.1–96.2) \\
 & Qwen3-30B-A3B-2507~\cite{yang2025qwen3} & 95.1 (93.9–96.1) & 94.4 (93.2–95.5) & 97.7 (96.8–98.3) \\
\addlinespace
\multirow{2}{*}{LumiOpen}
 & Llama-Poro-2-8B~\cite{poro2_2025}$^{\dagger}$ & 70.2 (67.1–73.0) & 69.3 (66.3–72.2) & 82.1 (79.5–84.5) \\
 & Llama-Poro-2-70B~\cite{poro2_2025}$^{\dagger}$ & 90.6 (88.5–92.4) & 91.8 (89.9–93.5) & 92.5 (90.6–94.1) \\
\addlinespace
Baseline & Rule-based (substring) & 45.4 (42.9–47.7) & 45.4 (42.9–47.7) & \textbf{100} (99.8–100.0) \\
\bottomrule
\end{tabular}
\end{table*}

\begin{table}[ht]
\centering
\footnotesize
\caption{\textbf{Pairwise McNemar tests comparing the best-performing model with competing models.}
The contingency counts report the number of questions where Model A was correct and Model B incorrect ($n_{10}$), and where Model A was incorrect and Model B correct ($n_{01}$). The McNemar test was computed using the exact binomial formulation on paired predictions across $N=1{,}664$ questions. Holm correction was applied for multiple comparisons.}
\label{tab:supp_mcnemar}
\begin{tabular}{llccc}
\toprule
\textbf{Model A} & \textbf{Model B} & $\boldsymbol{n_{10}}$ & $\boldsymbol{n_{01}}$ & \textbf{$p$-value} \\
\midrule
Llama-3.1-70B & Qwen3-30B-A3B-it-2507 & 35 & 31 & 0.843 \\
Llama-3.1-70B & R1-Distill-Llama-70B & 36 & 27 & 0.843 \\
Llama-3.1-70B & Gemma-3-27B-it & 70 & 27 & $<0.001$ \\
Llama-3.1-70B & MedGemma-27B-text-it & 83 & 30 & $<0.001$ \\
Llama-3.1-70B & Llama-3.3-70B & 19 & 12 & 0.843 \\
\bottomrule
\end{tabular}
\end{table}

\clearpage
\section*{H. Model Reproducibility}
\label{supp:reproducibility}

All instruction-tuned models evaluated in this study were obtained from their official Hugging Face repositories. To ensure exact reproducibility, we specify the repository identifier and the corresponding commit hash for each model, which uniquely defines the repository state from which model weights, configuration files, and tokenizers were loaded for inference. Table~\ref{supp:full_model_info} lists these identifiers for all models included in the experiments. No modifications were made to model architectures or weights prior to evaluation.

\begin{table*}[h!]
\centering
\footnotesize
\caption{Full reproducibility information for all instruction-tuned models evaluated in this study. Includes Hugging Face repository identifiers and exact commit hashes used for inference.}
\label{supp:full_model_info}
\begin{tabular}{ll}
\toprule
\textbf{Hugging Face Repository} & \textbf{Commit Hash} \\
\midrule
\texttt{google/gemma-3-4b-it} & 093f9f388b31de276ce2de164bdc2081324b9767 \\
\texttt{google/gemma-3-27b-it} & 005ad3404e59d6023443cb575daa05336842228a  \\
\texttt{google/medgemma-4b-it} & 290cda5eeccbee130f987c4ad74a59ae6f196408  \\
\texttt{google/medgemma-27b-text-it} & 5b667cf2ddcf064085bc90952edb35a0edbfb79c \\
\texttt{deepseek-ai/DeepSeek-R1-Distill-Qwen-7B} & 916b56a44061fd5cd7d6a8fb632557ed4f724f60 \\
\texttt{deepseek-ai/DeepSeek-R1-Distill-Qwen-32B} & 711ad2ea6aa40cfca18895e8aca02ab92df1a746  \\
\texttt{deepseek-ai/DeepSeek-R1-Distill-Llama-70B} & b1c0b44b4369b597ad119a196caf79a9c40e141e  \\
\texttt{LumiOpen/Llama-Poro-2-8B-Instruct} & dce5619131a6dd295633045e51d00c12e8b0133c \\
\texttt{LumiOpen/Llama-Poro-2-70B-Instruct} & ba7a467a544e2b8d944a8a8636120fd0fea9358d \\
\texttt{meta-llama/Llama-3.1-8B-Instruct} & 0e9e39f249a16976918f6564b8830bc894c89659  \\
\texttt{meta-llama/Llama-3.1-70B-Instruct} & 1605565b47bb9346c5515c34102e054115b4f98b \\
\texttt{meta-llama/Llama-3.3-70B-Instruct} & 6f6073b423013f6a7d4d9f39144961bfbfbc386b \\
\texttt{meta-llama/Llama-4-Scout-17B-16E-Instruct} & 92f3b1597a195b523d8d9e5700e57e4fbb8f20d3 \\
\texttt{openai/gpt-oss-20b} & 6cee5e81ee83917806bbde320786a8fb61efebee \\
\texttt{Qwen/Qwen3-4B-Instruct-2507} & cdbee75f17c01a7cc42f958dc650907174af0554 \\
\texttt{Qwen/Qwen3-30B-A3B-Instruct-2507} & 0d7cf23991f47feeb3a57ecb4c9cee8ea4a17bfe \\
\bottomrule
\end{tabular}
\end{table*}

\clearpage
\section*{I. GPU Memory Profiling}
\label{supp:memory}

Prior to the main experiments, memory profiling was conducted in an open high-performance computing environment using hardware with greater capacity than the deployment system in order to characterize memory requirements across a wide range of input lengths. Profiling was performed on three NVIDIA H200 GPUs, whereas the main experiments were restricted to a secure offline environment equipped with two NVIDIA A100 GPUs (80 GB each), providing a total memory budget of 160 GB. The objective was to determine which models could process long input sequences within this deployment constraint and to estimate the maximum feasible context length for each architecture.

Memory usage was measured as a function of input sequence length using synthetic inputs of predefined sizes. For a sequence of length $L$, the input consisted of a deterministic series of token identifiers $\{1,2,\ldots,L\}$, enabling precise control of input length without reliance on natural language content. GPU memory was recorded during a single forward decoding step in which the model processed the full input and computed next-token logits (equivalently, one-token generation). Model outputs were not analyzed, as the objective was solely to quantify memory requirements as a function of context length.

\begin{figure*}[h!]
    \centering
    \includegraphics[width=1\textwidth]{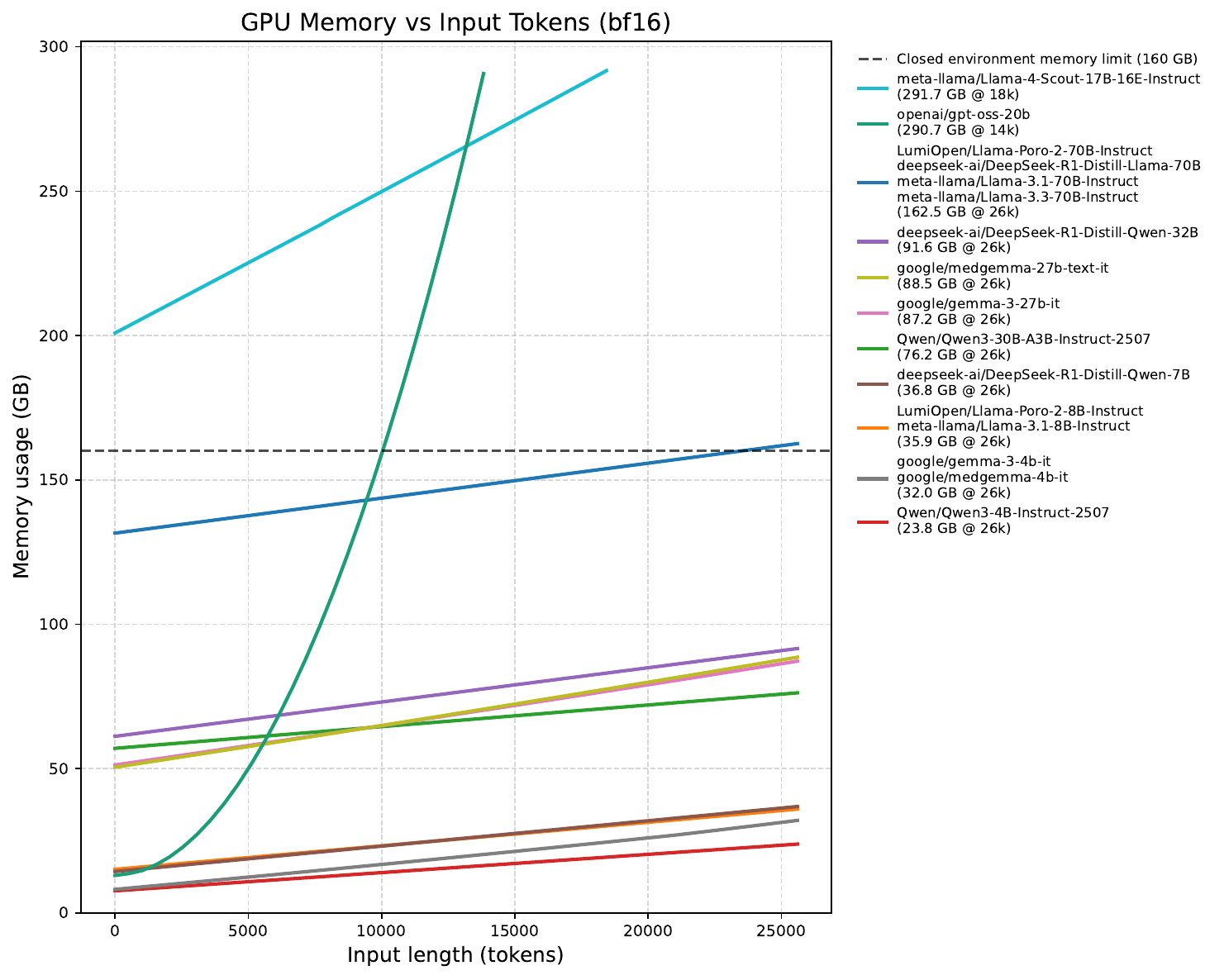}
    \caption{GPU memory usage as a function of input token length for bfloat16 precision. The dashed horizontal line indicates the 160~GB memory budget used in the main experiments.}
    \label{fig:memory_bf16}
\end{figure*}

\begin{figure*}[h!]
    \centering
    \includegraphics[width=1\textwidth]{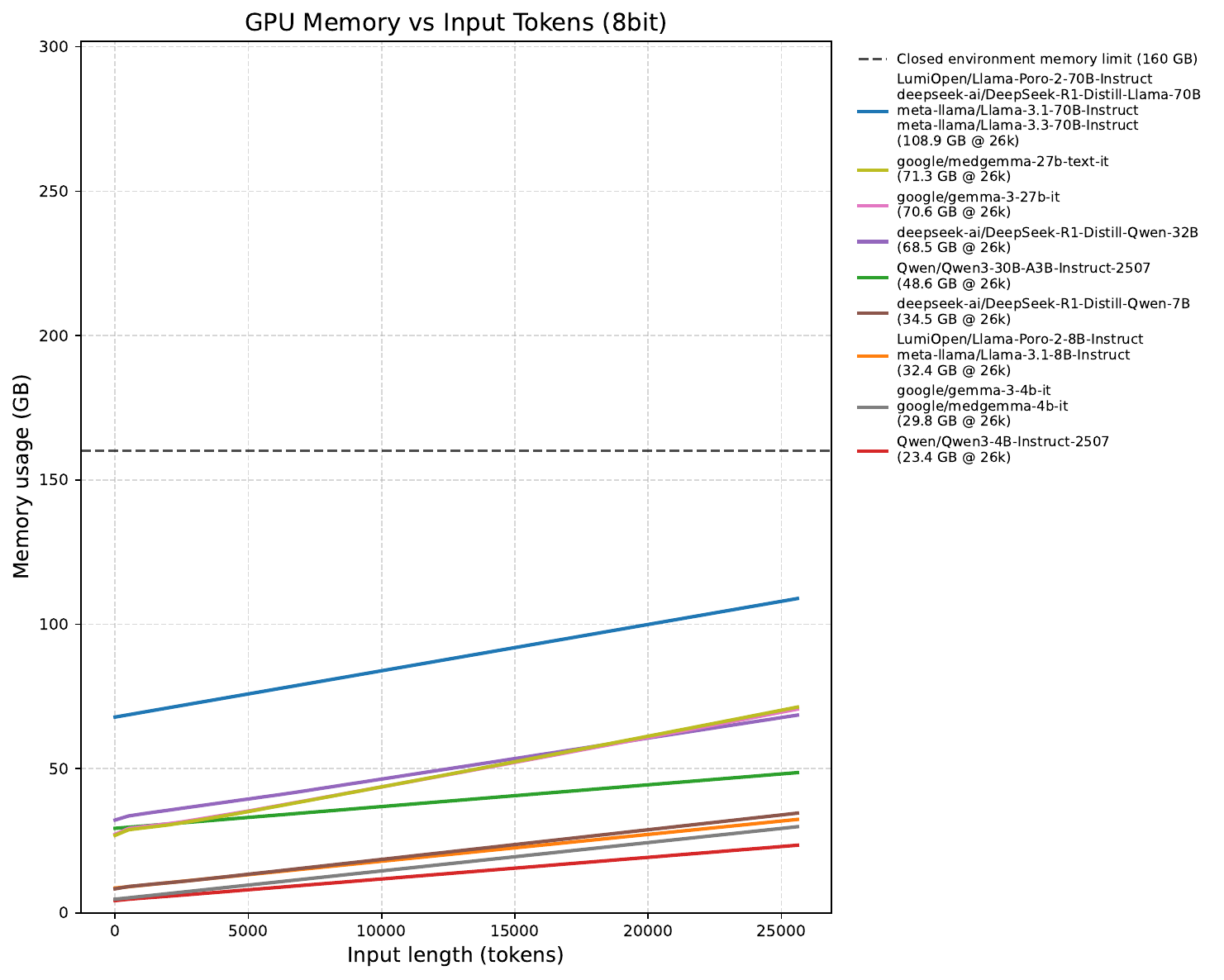}
    \caption{GPU memory usage as a function of input token length for 8-bit quantization. The dashed horizontal line indicates the 160~GB memory budget used in the main experiments. Llama-4 and GPT-OSS models are not included because these architectures are not compatible with bitsandbytes quantization.}
    \label{fig:memory_8bit}
\end{figure*}

\begin{figure*}[h!]
    \centering
    \includegraphics[width=1\textwidth]{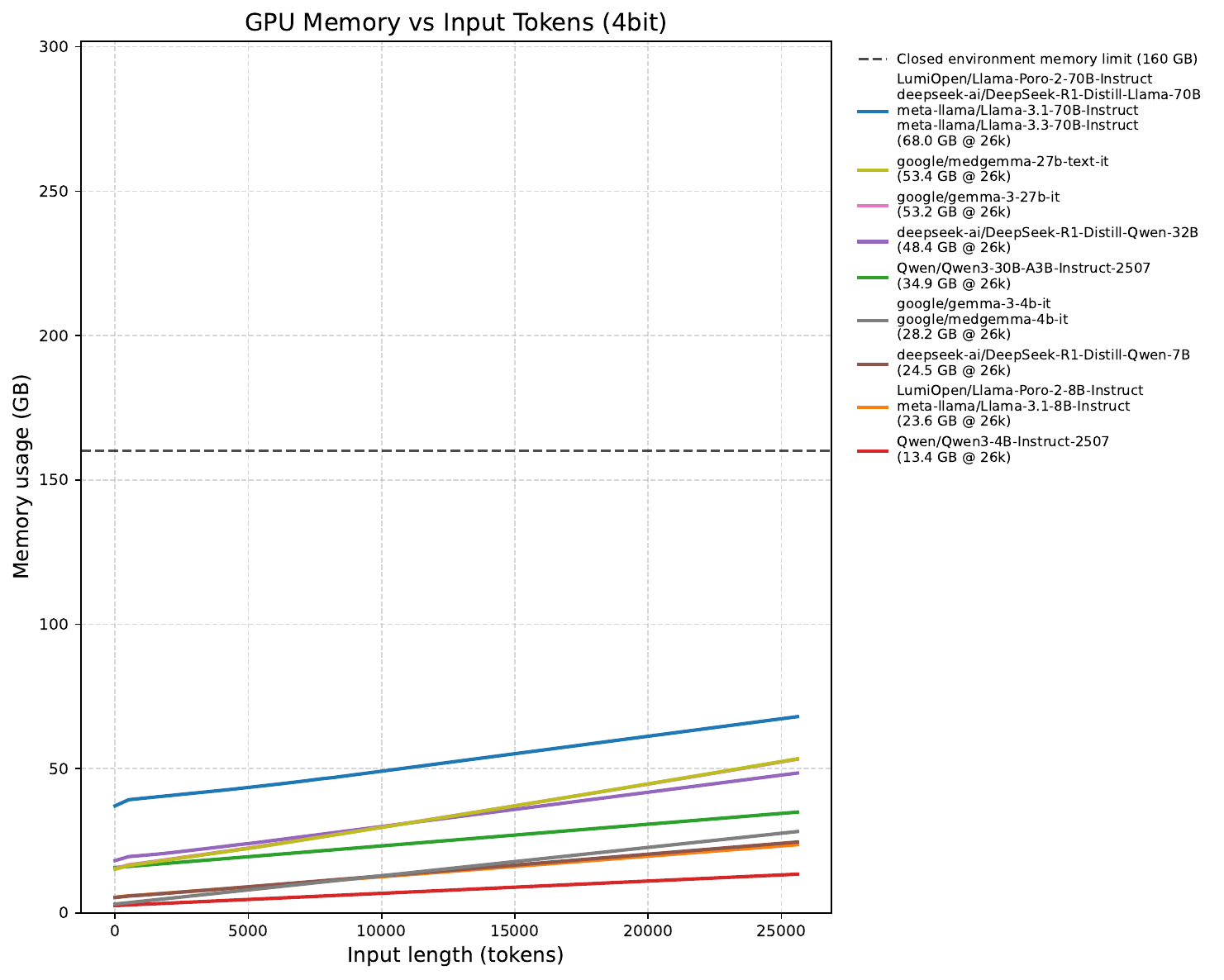}
    \caption{GPU memory usage as a function of input token length for 4-bit quantization. The dashed horizontal line indicates the 160~GB memory budget used in the main experiments. Llama-4 and GPT-OSS models are not included because these architectures are not compatible with bitsandbytes quantization.}
    \label{fig:memory_4bit}
\end{figure*}

Sequence lengths ranged from 1 to 25{,}120 tokens. Lengths of 512 tokens and above were evaluated in increments of 512 tokens, with an additional single-token input included to capture the minimal context case. Figures~\ref{fig:memory_bf16}--\ref{fig:memory_4bit} report memory usage for bfloat16 precision and for 8-bit and 4-bit quantized models, respectively. In each plot, the dashed horizontal line denotes the 160~GB memory budget applied in the main experiments.

Profiling was performed in a non-clinical environment without patient data, allowing unrestricted scaling of input length. Each model was profiled in a separate isolated Python process to ensure a clean memory state and to avoid cross-run memory retention. Within each model run, sequence lengths were evaluated sequentially from shortest to longest while tracking peak GPU memory usage after each inference step.

Results for 4-bit and 8-bit quantization do not include the \texttt{meta-llama/Llama-4-Scout-17B-16E-Instruct} and \texttt{openai/gpt-oss-20b} models, as these architectures are not compatible with the bitsandbytes quantization library used in this study.

\end{appendices}

\clearpage
\bibliography{sn-bibliography}

\end{document}